\documentclass[a4paper,fleqn]{cas-dc}

\usepackage[authoryear]{natbib}

\usepackage[ruled,linesnumbered]{algorithm2e}
\usepackage{setspace}

\usepackage{graphicx}
\usepackage{subfig}

\usepackage{makecell}

\usepackage{soul}

\usepackage{tcolorbox}
\usepackage[dvipsnames]{xcolor}

\usepackage{pifont}

\usepackage{hyperref}
\hypersetup{
    colorlinks=true,
    linkcolor=Blue, 
    urlcolor=Blue,  
    citecolor=Blue, 
}

\usepackage{caption}
\def\tsc#1{\csdef{#1}{\textsc{\lowercase{#1}}\xspace}}
\tsc{WGM}
\tsc{QE}

\begin{document}
\let\WriteBookmarks\relax
\def\floatpagepagefraction{1}
\def\textpagefraction{.001}

\shorttitle{<OkayPlan>}    

\shortauthors{<Xin et al.>}  

\title [mode = title]{\textbf{OkayPlan}: \textbf{O}bstacle \textbf{K}inematics \textbf{A}ugmented D\textbf{y}namic Real-time Path \textbf{Plan}ning via Particle Swarm Optimization}

\author[1]{Jinghao Xin}[orcid=0000-0001-9875-9098]
\ead{xjhzsj2019@sjtu.edu.cn}
\ead[url]{https://github.com/XinJingHao}
\credit{Conceptualization, Methodology, Software, Investigation, Writing - Original Draft}

\author[2,3]{Jinwoo Kim}[orcid=0000-0002-2237-4965]
\ead{jinwookim@hanyang.ac.kr}
\ead[url]{https://sites.google.com/view/jinwoo-kim}
\credit{Resources, Supervision, Funding acquisition}

\author[1]{Shengjia Chu}[orcid=0000-0003-1390-8512]
\ead{chu-sjtu@sjtu.edu.cn}
\credit{Validation, Writing - Review \& Editing}

\author[1]{Ning Li}[orcid=0000-0003-1025-9641]
\cormark[1] 
\ead{ning_li@sjtu.edu.cn}
\ead[url]{https://academic532.github.io}
\credit{Resources, Supervision, Funding acquisition}

\cortext[1]{Corresponding author}

\address[1]{Department of Automation, Shanghai Jiao Tong University, Shanghai 200240, P.R. China}
\address[2]{School of Civil and Environmental Engineering, Nanyang Technological University, S639798, Singapore}
\address[3]{Department of Civil and Environmental Engineering, Hanyang University, Seoul 04763, Republic of Korea}

\begin{abstract}
Existing Global Path Planning (GPP) algorithms predominantly presume planning in static environments. This assumption immensely limits their applications to Unmanned Surface Vehicles (USVs) that typically navigate in dynamic environments. To address this limitation, we present OkayPlan, a GPP algorithm capable of generating safe and short paths in dynamic scenarios at a real-time executing speed (125 Hz on a desktop-class computer). Specifically, we approach the challenge of dynamic obstacle avoidance by formulating the path planning problem as an Obstacle Kinematics Augmented Optimization Problem (OKAOP), which can be efficiently resolved through a PSO-based optimizer at a real-time speed. Meanwhile, a Dynamic Prioritized Initialization (DPI) mechanism that adaptively initializes potential solutions for the optimization problem is established to further ameliorate the solution quality. Additionally, a relaxation strategy that facilitates the autonomous tuning of OkayPlan's hyperparameters in dynamic environments is devised. Comprehensive experiments, including comparative evaluations, ablation studies, and \textcolor{black}{applications to 3D physical simulation platforms}, have been conducted to substantiate the efficacy of our approach. Results indicate that OkayPlan outstrips existing methods in terms of path safety, length optimality, and computational efficiency, establishing it as a potent GPP technique for dynamic environments. The video and code associated with this paper are accessible at \url{https://github.com/XinJingHao/OkayPlan}.
\end{abstract}


\maketitle

\section{Introduction}
Unmanned Surface Vehicles (USVs), a novel class of maritime vessels endowed with advanced capabilities such as autonomous perception, navigation, and mission execution, have been widely employed in various domains, including ocean resource exploration \citep{app_ore}, maritime traffic monitoring \citep{app_mtm}, environmental protection \citep{app_ep}, and water rescue \citep{app_wr}. As a critical component of USVs, the path planning techniques have underpinned their autonomous navigation capabilities and enabled the execution of unmanned tasks, which have received considerable attention in recent years \citep{OE_USV_DDPG, OE_USV_IBA, OE_MultiUSV}.

Path planning is typically bifurcated into two categories: Global Path Planning (GPP) and Local Path Planning (LPP). The GPP algorithm is tasked with generating a collision-free global path with minimal length given the environmental map, and the LPP algorithm governs the USVs in adhering to the global path while concurrently circumventing the newly emerged obstacles. The global path functions as a higher-level instruction for the LPP algorithm, underlying the autonomous navigation of USVs. Consequently, the development of advanced GPP algorithms is of paramount importance and thus constitutes the main focus of this research. 

\textcolor{black}{Canonical GPP algorithms encompass three primary classes: search-based, sample-based, and metaheuristic-based. Search-based GPP algorithms, a type of approach that searches for viable paths by exploring nodes within graph-based or grid-based maps, can be traced back to the foundational Breadth-First Search \citep[BFS;][]{BFS}. BFS explores nodes equidistant from the starting node stage-by-stage until a viable path is identified. This simple mechanism ensures the discovery of the shortest path in unweighted maps. Based on this pioneering work, extensive methods have been proposed subsequently, including Dijkstra \citep{Dijkstra}, A* \citep{Astar}, Bidirectional A* \citep{BAstar}, Theta* \citep{theta_star}, Jump Point Search \citep[JPS;][]{JPS}, and Improved Bidirectional A* \citep[IBA*;][]{OE_USV_IBA}. These works primarily focus on (i) extending the methodology to weighted maps and (ii) reducing the search space to improve computational efficiency. Sample-based GPP algorithms, such as Probabilistic Roadmaps \citep[PRM;][]{PRM} and Rapidly-exploring Random Trees \citep[RRT;][]{RRT}, construct feasible paths via sampling directly in the configured space. This typically renders them computationally faster than search-based methods. However, this advantage comes at the cost of compromising path smoothness and length optimality. To address this issue, various improved methods have been developed, including RRT* \citep{RRTstar}, RRT*-Smart \citep{RRTstar_smart}, Informed RRT* \citep{IRRTstar}, Expanding Path RRT* \citep{EP_RRTstar}, and Plant Grow Route \citep[PGR;][]{PGR}. Regarding the metaheuristic-based GPP algorithms, the path planning problem is formulated as an optimization problem, with the goal of minimizing the path length, while collision constraints are incorporated as penalty terms. Due to the non-differentiability of the optimization problem, gradient-free metaheuristic optimizers, such as Genetic Algorithm \citep[GA;][]{GA} and Particle Swarm Optimization \citep[PSO;][]{PSO}, are resorted. Owing to their simple concept and easy implementation, metaheuristic-based GPP algorithms have gained widespread adoptions in the domains of mobile robots \citep{PSO_mobilerobot}, Unmanned Aerial Vehicles \citep[UAVs;][]{PSO_UAV}, and USVs \citep{PSO_USV}. Conventionally, metaheuristic-based planners involve iterating over a large number of candidate solutions, often resulting in prolonged planning time. To mitigate this drawback, parallel computation techniques have been introduced. Representative methods include: Parallel Genetic Algorithm \citep[PGA;][]{PGA}, Diversity-based Parallel Particle Swarm Optimization \citep[DPPSO;][]{DPPSO}, and Self-Evolving Particle Swarm Optimization \citep[SEPSO;][]{SEPSO}.}

\begin{figure*}[t]
\centering
\includegraphics[width=0.95\textwidth]{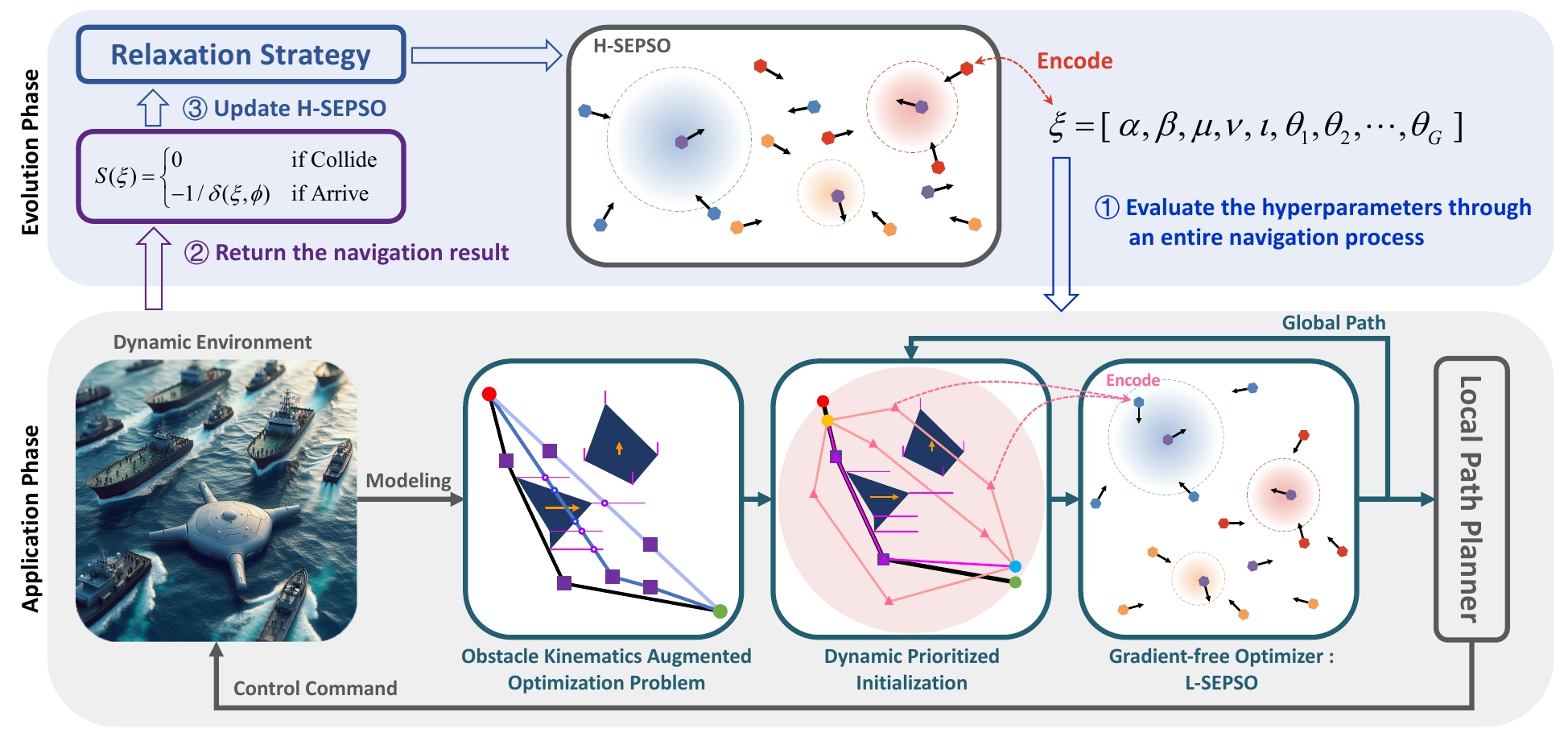}
\caption{\textcolor{black}{An overview of OkayPlan.}}
\label{pipline}
\end{figure*}

\textcolor{black}{Despite extensive research efforts, a significant gap persists in developing highly efficient GPP algorithms for USVs. Firstly, a common limitation of prior research is the failure to accommodate the motion of dynamic obstacles in GPP, instead delegating obstacle avoidance solely to LPP. While suitable for well-controlled systems like mobile robots, this approach is irrational for USVs facing challenges of underactuated control \citep{USV_underact}, nonlinear dynamics \citep{USV_nonlinear}, environmental disturbances \citep{VRX_ctrl}, and uncertainties \citep{USV_uncertainties}. These challenges characterize USVs with inaccurate and delayed maneuverability, risking collision when avoiding dynamic obstacles with merely a shortsighted LPP. Hence, incorporating the motion of dynamic obstacles in GPP and addressing potential collisions from a global perspective could significantly enhance USVs' safety. Secondly, few of these GPP algorithms are capable of achieving real-time planning. For instance, PGA \citep{PGA}: 3.63 \textit{s/p} (seconds per planning); PGR \citep{PGR}: 6.5$\sim$6.9 \textit{\textit{s/p}}; IBA* \citep{OE_USV_IBA}: 2.568 \textit{s/p}. As USVs typically operate in dynamic marine environments with disturbances from wind, waves, and other vessels, the ability to supplement pre-planned paths with real-time adjustments is indispensable, necessitating the design of computationally efficient GPP algorithms. Lastly, given USVs' limited onboard energy resources, identifying shorter paths in dynamic environments could significantly extend their operational range.}

\textcolor{black}{To address the aforementioned limitations, we have developed OkayPlan, an improved metaheuristic-based global path planning algorithm tailored for dynamic marine environments. OkayPlan is capable of generating safer and shorter paths while achieving real-time planning at a high frequency of 125 Hz (approximately 0.008 \textit{s/p}).} The key contributions that underprop the OkayPlan are summarized as follows:

\begin{itemize}
\item An Obstacle Kinematics Augmented Optimization Problem (OKAOP) has been formulated, allowing the motion of dynamic obstacles to be seamlessly incorporated, thereby facilitating the derivation of safer paths.

\item A Dynamic Prioritized Initialization (DPI) mechanism has been introduced. The DPI adaptively initializes the particles of OkayPlan in accordance with different planning stages, which bolsters its optimization capability and contributes to the rapid identification of shorter paths in dynamic environments.

\item A relaxation strategy has been established to subdue the stochasticity of the dynamic environments, enabling a better tuning of OkayPlan's hyperparameters.
\end{itemize}

The remainder of the paper is structured as follows: Section 2 reviews the GPP optimization problem and prior research on OkayPlan. Section 3 introduces the OkayPlan algorithm. Section 4 presents experimental results. Finally, Section 5 concludes the paper and outlines future research directions.
\section{Preliminaries}

\subsection{Problem formulation of global path planning}
\begin{figure}[t]
\centering
\includegraphics[width=0.4\textwidth]{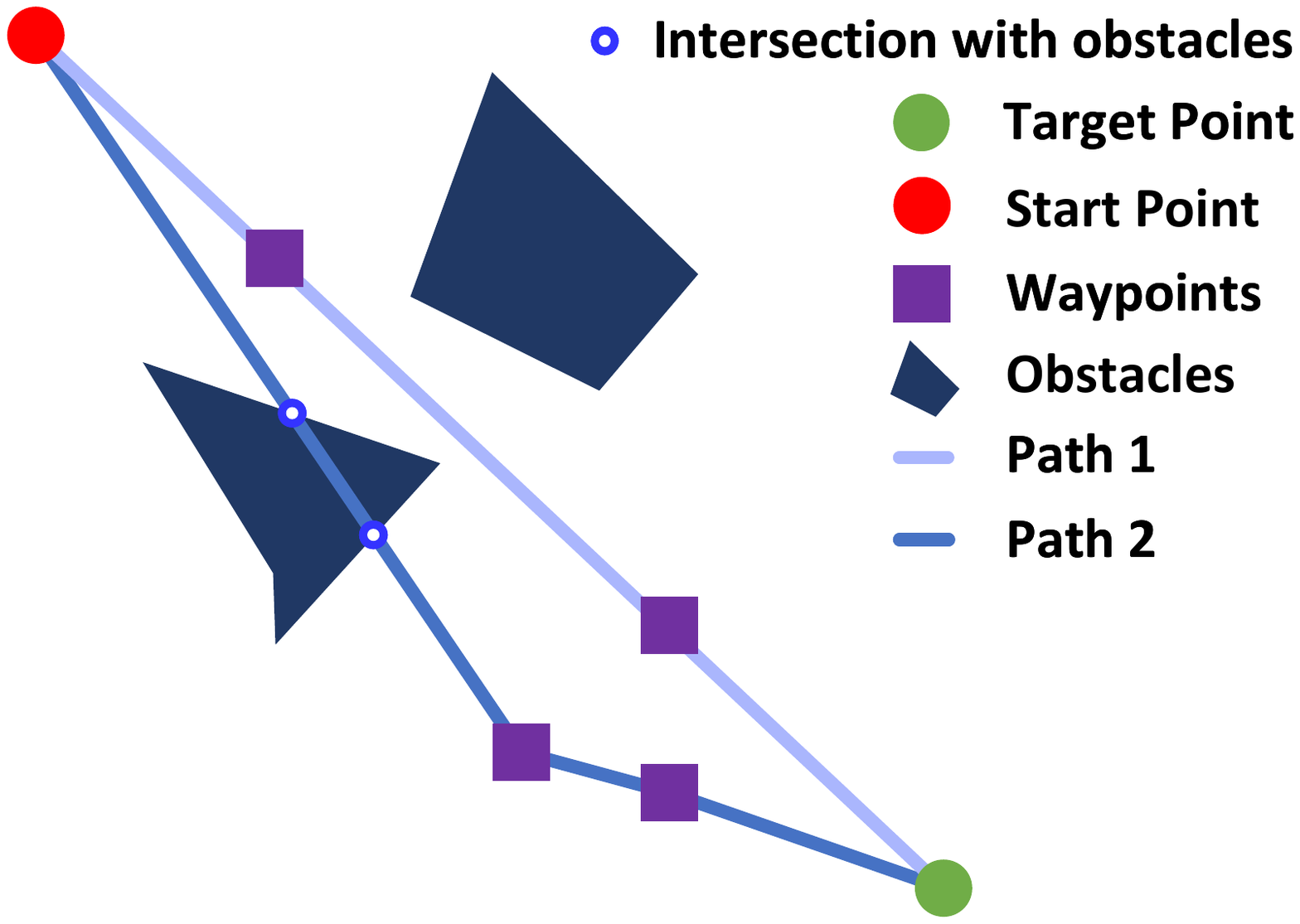}
\caption{Problem formulation of global path planning.}
\label{GPP}
\end{figure}

As introduced by \cite{SEPSO}, the 2D GPP problem can be formulated as an optimization problem illustrated in Fig. \ref{GPP}, wherein the decision variables are the coordinates of the waypoints from a specific path, as denoted below:
\begin{equation}
\label{coordinate}
Z=[x_1,x_2,...,x_{D/2},y_1,y_2,...,y_{D/2}]
\end{equation}

\noindent where $D/2$ corresponds to the number of the waypoints of each path, with $D$ denoting the dimension of the search space. The objective function of the optimization problem is given below:
\begin{equation}
\label{obj}
F(Z)=L(Z)+\alpha \cdot Q(Z)^\beta
\end{equation}

\begin{equation}
\begin{aligned}
\label{lenth}
L(Z)&=\sum^{D/2}_{d=2}\sqrt{(x_d-x_{d-1})^2+(y_d-y_{d-1})^2}\\
&+\sqrt{(x_t-x_{D/2})^2+(y_t-y_{D/2})^2}\\
&+\sqrt{(x_s-x_{1})^2+(y_s-y_{1})^2}
\end{aligned}
\end{equation}

\noindent where $L(Z)$ denotes the path length; $(x_t, y_t)$ and $(x_s, y_s)$ are the coordinates of the target and start point; the parameters $\alpha$ and $\beta$ govern the magnitude of the penalty incurred from collision, with $Q(Z)$ denoting the count of intersections between the path and the bounding boxes of the obstacles. As shown in Fig. \ref{GPP},  $Q(\text{Path 1})=0$, and $Q(\text{Path 2})=2$.

Though minimizing Eq. (\ref{obj}), one is expecting to identify a collision-free path with minimal length.

\subsection{Self-evolving particle swarm optimization}\label{SEPSO_preliminary}
Due to the non-differentiability of Eq. (\ref{obj}), gradient-based optimization methods are inapplicable. Although gradient-free optimization techniques such as GA and PSO are generally feasible, their unsatisfactory computational efficiency and susceptibility to local optima render them less than ideal for the GPP problem. Extensive attempts have been undertaken to alleviate this limitation in recent decades \citep{IPSO1,IPSO2,DPPSO}. Among these, DPPSO \citep{DPPSO} has drawn much attention due to its simple concept and robust performance. In DPPSO, the standard PSO is split into multiple groups, each of which performs diversified search parameters $\theta_g$ to preserve population diversity and thus avoid local optima. Meanwhile, the information of different groups is shared to ensure convergence. The pivotal update formulas of DPPSO are as follows:
\begin{equation}
\begin{aligned}
\label{DPPSO_V}
V_{g,n}^{k+1}=\omega^{k}_{g} V_{g,n}^{k}&+C1_{g} \cdot R1^k_{g,n}  \cdot\left(\mathrm{Pbest}_{g,n}^{k}-X_{g,n}^{k}\right) \\
&+C2_{g} \cdot R2^k_{g,n} \cdot\left(\mathrm{Gbest}^{k}_{g}-X_{g,n}^{k}\right) \\
&+C3_{g} \cdot R3^k_{g,n} \cdot\left(\mathrm{Tbest}^{k}-X_{g,n}^{k}\right) 
\end{aligned}		
\end{equation}

\begin{equation}	
\label{DPPSO_X}
X_{g,n}^{k+1}=X_{g,n}^{k}+V_{g,n}^{k+1}	
\end{equation}

\begin{equation}
\label{omega}
\omega_{g}^{k}=\omega_{g}^{init}-\frac{\omega_{g}^{init}-\omega_{g}^{end}}{T} \cdot k
\end{equation}

\begin{equation}
\label{params_theta}
\theta_{g}=[C1_g, C2_g, C3_g, \omega_{g}^{init}, \omega_{g}^{end}, V_{g}^{limit}]
\end{equation}

\noindent where $k$ is the iteration counter; $T$ is the maximum iteration times; $g$ is the group's index within the total group number $G$; $n$ is the particle's index within its group, with a maximum value of $N$; $V_{g,n}^{k}$ and $X_{g,n}^{k}$ are the velocity and position of the particle $n$ in group $g$ at iteration $k$, respectively; $R1^k_{g,n}, R2^k_{g,n}, R3^k_{g,n}$ are three uniform random variables within range (0,1); $C1_g, C2_g, C3_g$ are three positive acceleration constants; $\omega_g^k$ is the inertia factor, with its initial and final value respectively given by $\omega_g^{init}$ and $\omega_g^{end}$; $V_g^{limit}$ is the hyperparameter that control the range of the velocity of group $g$ such that ${V_g^{range} = V_g^{limit} \cdot X^{range}}$, where $X^{range}$ is the search space of the optimization problem being considered; ${Pbest}_{g,n}^{k}$ is the best solution identified by the particle $n$ in group $g$ within iteration $k$; ${Gbest}_{g}^{k}$ is the best solution found by group $g$ within iteration $k$; ${Tbest}^{k}$ is the best solution discovered by the whole population within iteration $k$; $Pbest_{g,n}^k, Gbest_{g}^k, Tbest^k$ are updated according to Algorithm \ref{best_principle} at each iteration.

\begin{algorithm}[t]
\setstretch{1.2}
\SetAlgoNlRelativeSize{0}
\caption{Updating principle for the best values and positions}
\label{best_principle}
\KwIn{particle $X_{g,n}^k$}
\KwOut{the best positions $Pbest_{g,n}^k, Gbest_{g}^k, Tbest^k$ and the best values $U^{Pbest}_{g,n}, U^{Gbest}_{g}, U^{Tbest}$}
\tcp{$f(\cdot)$ is the fitness function}
\tcp{For minimization problems:}
\If{$f(X_{g,n}^k)<U^{Pbest}_{g,n}$}{$U^{Pbest}_{g,n} \leftarrow f(X_{g,n}^k)$ and $Pbest_{g,n}^k \leftarrow X_{g,n}^k$} 

\If{$f(X_{g,n}^k)<U^{Gbest}_{g}$}{$U^{Gbest}_{g} \leftarrow f(X_{g,n}^k)$ and $Gbest_{g}^k \leftarrow X_{g,n}^k$}

\If{$f(X_{g,n}^k)<U^{Tbest}$}{$U^{Tbest} \leftarrow f(X_{g,n}^k)$ and $Tbest^k \leftarrow X_{g,n}^k$}
\end{algorithm}

Through the above iteration, the particles dynamically explore the search space and progressively converge towards potential optima. Furthermore, the diversified search parameters employed by DPPSO attain a more favorable balance between exploration and exploitation, consequently leading to improved optimization performance. Nonetheless, the DPPSO encounters two noteworthy unresolved challenges: 1) its calculation speed is prohibitively slow due to its particle-wise manipulation; 2) determining the diversified search parameter, denoted as $\theta$ (a collection of search parameters from all groups, such that ${\theta=[\theta_1, \theta_2, \cdots \theta_g]}$), poses a significant difficulty. 

\begin{figure}[t]
\centering
\includegraphics[width=0.325\textwidth]{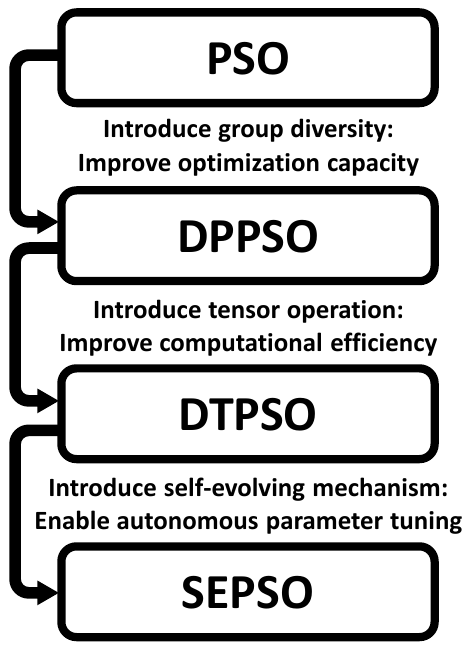}
\caption{\textcolor{black}{The evolution from PSO to SEPSO.}}
\label{PSO_evolution}
\end{figure}

To address these two issues, \cite{SEPSO} have introduced the SEPSO, which augmented the DPPSO with a TOF and a Hierarchical Self-Evolving Framework (HSEF). \textcolor{black}{The evolution of these algorithms is summarized in Fig. \ref{PSO_evolution}}. The TOF transforms the particle-wise manipulation into tensor operation, which endows the DPPSO with the feasibility to be deployed on the Graphics Processing Unit (GPU), thereby enabling parallel and rapid computation. Meanwhile, the HSEF facilitates autonomous tuning of the diversified search parameter $\theta$. Specifically, the HSEF consists of a Higher-level Self-Evolving Particle Swarm Optimization (H-SEPSO) and a Lower-level Self-Evolving Particle Swarm Optimization (L-SEPSO). Both the H-SEPSO and the L-SEPSO are underpinned by the TOF-augmented DPPSO \citep[DTPSO;][]{SEPSO}, each characterized by a distinct definition of particles and fitness functions. The L-SEPSO directly addresses the GPP problem, where each particle represents a potential path, and its fitness function corresponds to Eq. (\ref{obj}). Apparently, a lower fitness value indicates a more favorable path. Recall that the optimization capability of the L-SEPSO is notably conditioned by its search parameter $\theta$. Then, the L-SEPSO can be conceptualized as a mapping from $\theta$ to the Lowest Fitness Value (LFV) it could optimize: 
\begin{equation}
\label{LFV}
\text{LFV} = \text{L-SEPSO}(\theta)
\end{equation}

Minimizing Eq. (\ref{LFV}) holds the potential to identify a search parameter with better optimization capability. This optimization process is actually executed by the H-SEPSO, whose particle corresponds to the search parameter $\theta$ and fitness function aligns with Eq. (\ref{LFV}). Though alternatively conducting the optimization of the L-SEPSO and H-SEPSO, $\theta$ autonomously evolves by itself and converges toward a more rational configuration ultimately. This entire process is referred to as the evolution phase of SEPSO. Note that the evolution phase is computationally demanding, underscoring the imperative role of the efficient TOF.

\subsection{Global path planning with SEPSO and prioritized initialization}\label{Chapter_PI}

\begin{figure}[b]
\centering
\includegraphics[width=0.45\textwidth]{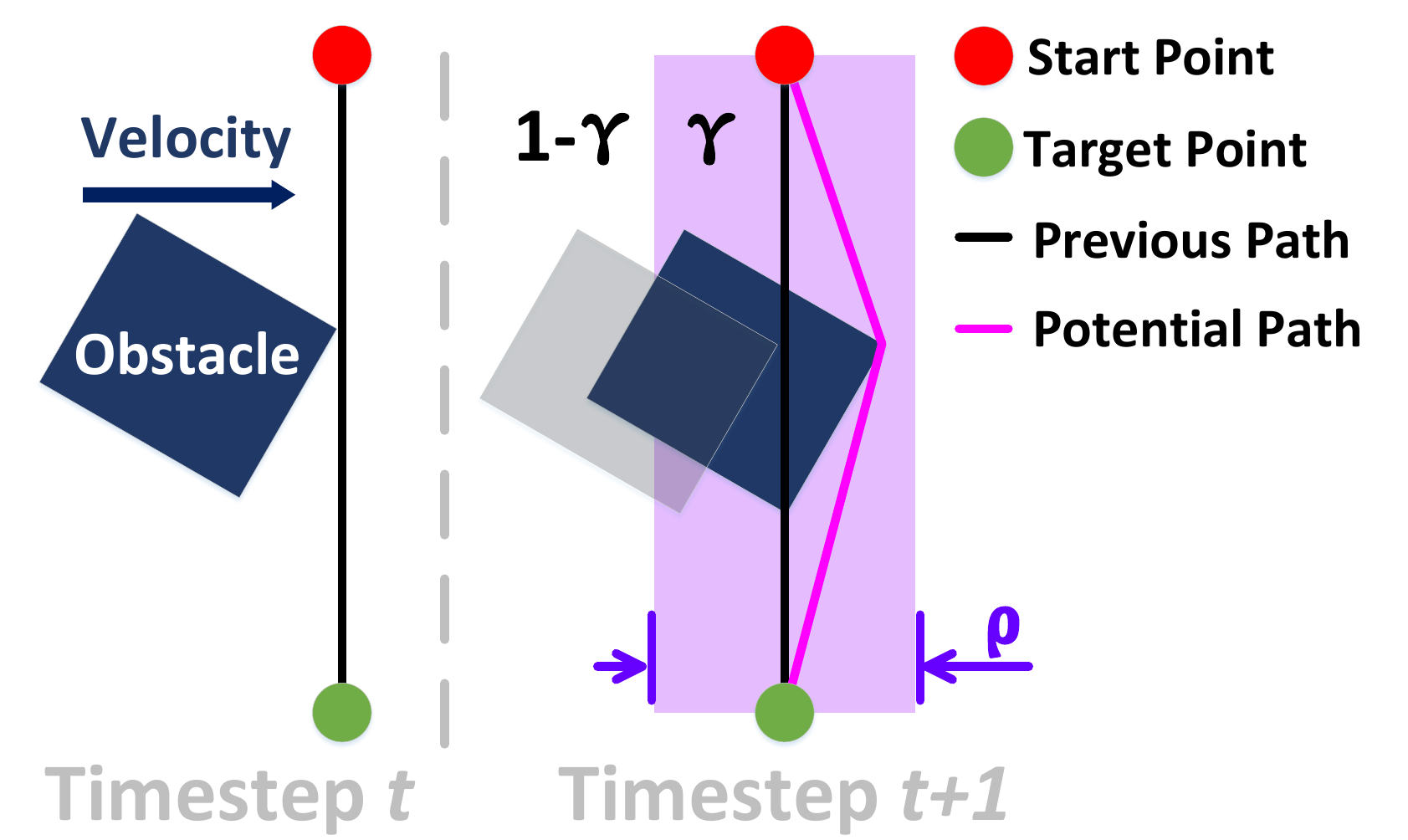}
\caption{Prioritized Initialization Mechanism.}
\label{PI}
\end{figure}

Following the evolution phase outlined in the last section, the refined search parameter $\theta$ is then applied in conjunction with the DTPSO to tackle the GPP problem formulated by Eq. (\ref{obj}), which is denoted as the application phase of the SEPSO. As widely recognized, the PSO-based optimization techniques are exceptionally suspicious to the initialization of the particles. A deliberate initialization serves not only to enhance solution quality but also to expedite convergence. In consideration of this, \cite{SEPSO} have devised a Prioritized Initialization (PI) mechanism to assist particle initialization. As depicted by Fig. \ref{PI}, at the commencement of each planning timestep, the PI mechanism inherits the planned path from the last timestep and randomly initializes a fraction of the population,  denoted by $\gamma$, within the PI interval $\rho$. This strategy sets the preceding planned path as a reference for the current timestep, thereby averting the whole population from searching from scratch. The insights behind the PI mechanism pertain to the continuous change of the environment, implying that the optimal paths of two consecutive planning timesteps are likely to manifest gradual variations rather than abrupt change in most cases. In addition, to ensure a judicious level of exploration, the PI mechanism stochastically initializes the remaining population, represented by $1-\gamma$, across the entire search space.

\section{Methodologies}
Having comprehended the fundamental knowledge associated with the GPP and SEPSO, we proceed to introduce the OkayPlan. We commence by introducing the OKAOP, wherein the motion of the obstacles is factored to derive safer paths. Following this, we propose a refinement of the PI, namely DPI, to initialize the population of SEPSO at different planning stages adaptively. Lastly, a relaxation strategy for the evolution phase of the SEPSO has been developed, aiming to handle the stochasticity stemming from the dynamic environments.
\subsection{Obstacle kinematics augmented optimization problem}\label{Chaper:OKA}

\begin{figure}[t]
\centering
\includegraphics[width=0.43\textwidth]{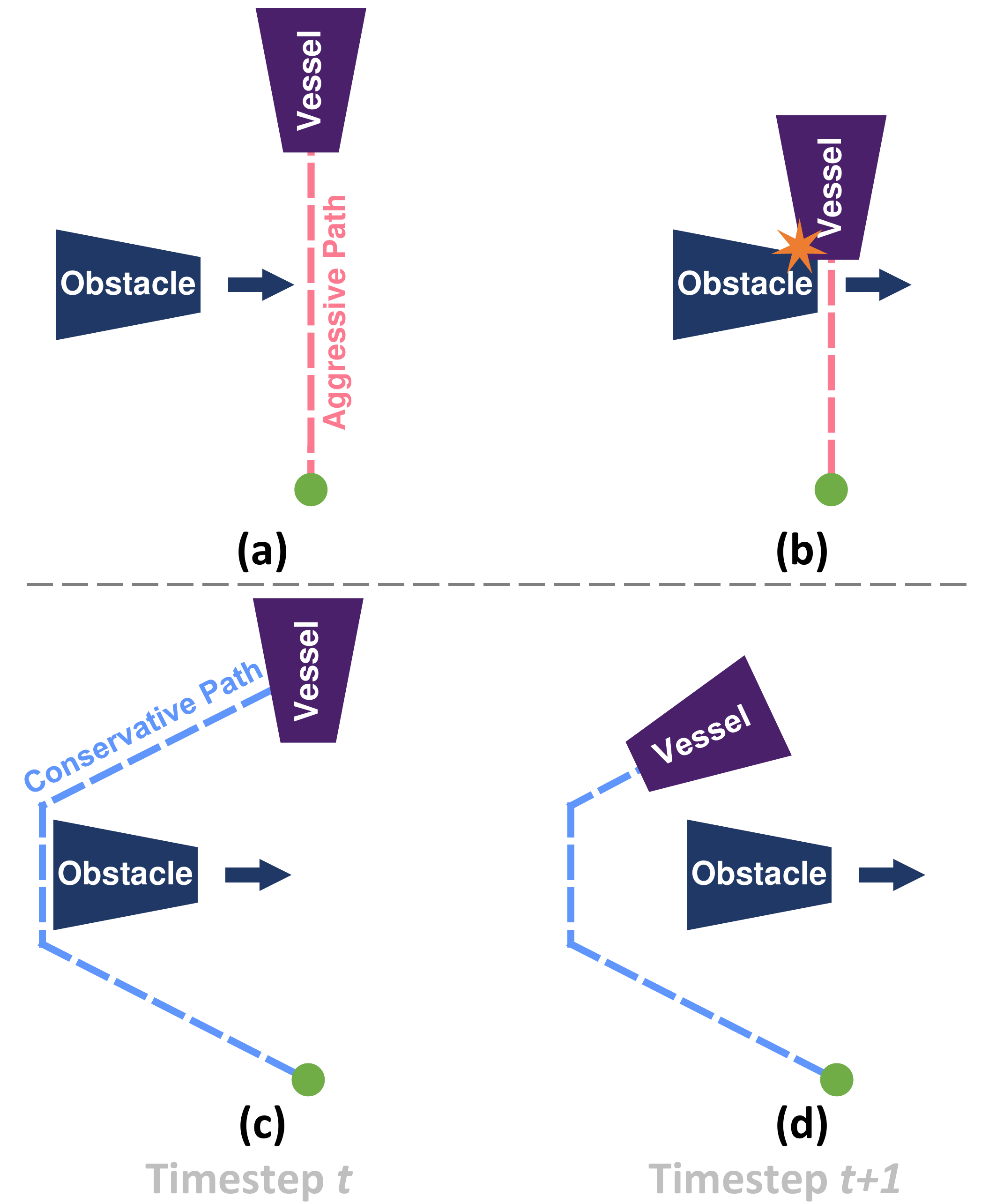}
\caption{A comparison of aggressive and conservative path planning.}
\label{AC}
\end{figure}

As illustrated by Fig. \ref{AC}, a crucial factor in guaranteeing safe navigation within dynamic environments is the prevention of "contention" originating from aggressive path planning. In particular, Fig. \ref{AC} (a) and (b) reveal that while the aggressive path is optimal in terms of path length, it may not be the safest choice and could lead to collision when navigating in dynamic environments. A more judicious way is to bypass the obstacles in a conservative fashion, as exemplified in Fig. \ref{AC} (c) and (d). This observation has uncovered a contradiction with the GPP's objective function delineated by Eq. (\ref{obj}), since we are seeking to ascertain a collision-free path of minimal length, but a newly emerged question pertains to how and to what extent the length should be shortened. 

\begin{figure}[t]
\centering
\includegraphics[width=0.43\textwidth]{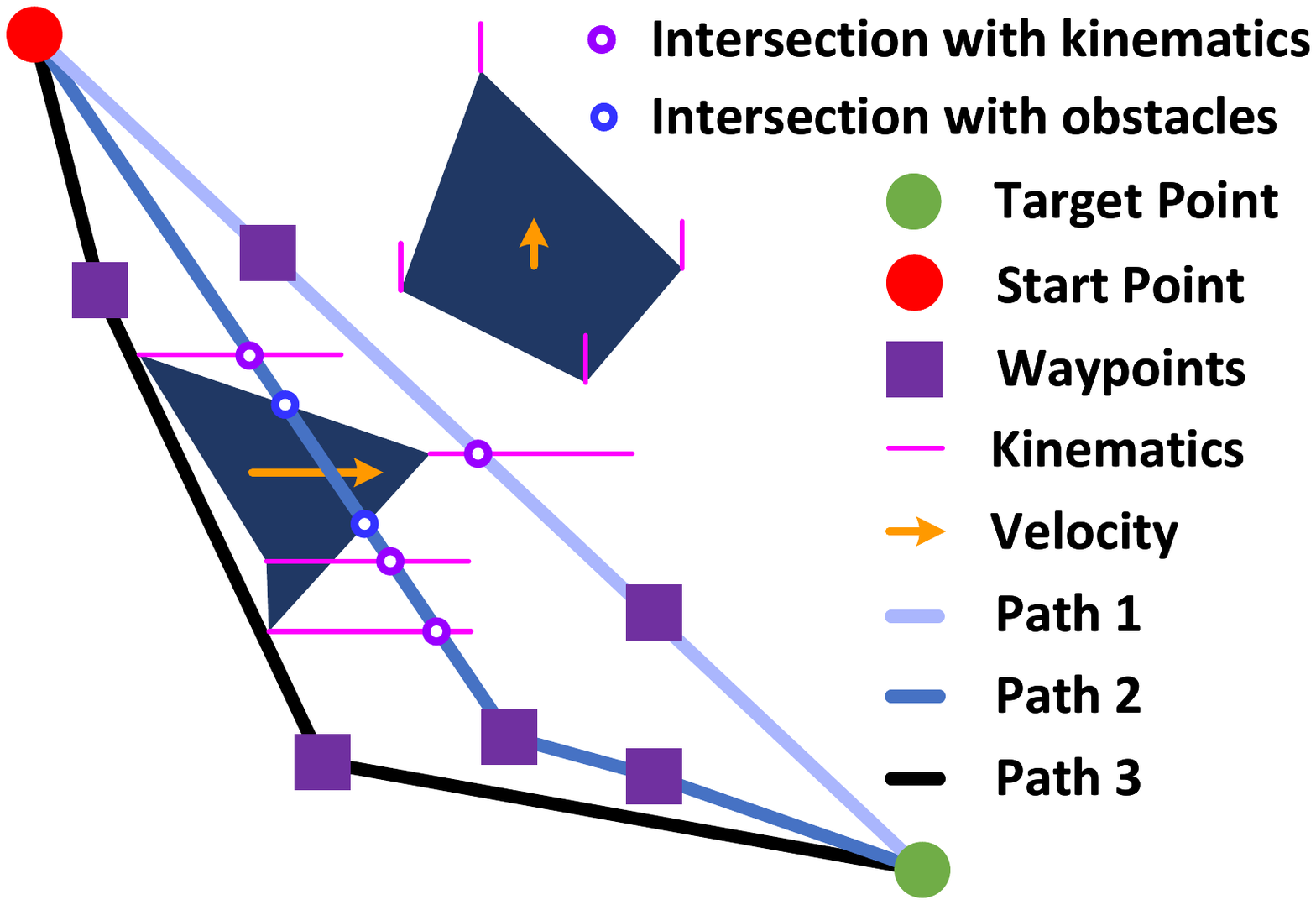}
\caption{\textcolor{black}{Obstacle kinematics augmented GPP in dynamic scenarios.}}
\label{OKA}
\end{figure}

We endeavor to address this issue by incorporating the obstacle kinematics into the objective function. To be more specific, we measure the instantaneous velocities of the obstacles, represent them with segments, and integrate them into the configured space, as the purple segments shown in Fig. \ref{OKA}. The obstacle kinematics can be construed as a rough prediction of the obstacle's future trajectory and be harnessed to assist the path planning. In order to ensure compatibility with the configured space, the length of the kinematic segments will be scaled by a hyperparameter, denoted as $\iota$, in proportion to the instantaneous velocities. In this context, the fitness function of OKAOP for GPP in dynamic environments is given by:
\begin{equation}
\label{dobj}
F(Z)=L(Z)+\alpha \cdot Q(Z)^\beta + \mu \cdot P(Z)^\nu
\end{equation}

\noindent where $\mu$ and $\nu$ are two hyperparameters that determine the magnitude of the penalty incurred from aggressive path planning, with $P(Z)$ denoting the count of intersections between the path and kinematic segments. As shown in Fig. \ref{OKA}, $P(\text{Path 1})=1$, $P(\text{Path 2})=3$, and $P(\text{Path 3})=0$. \textcolor{black}{Here, Path 2 is deemed the least desirable among the three routes, primarily attributable to its intersections with both obstacles and kinematic segments. Despite being the shortest, Path 1 incurs a penalty owing to its interaction with kinematic segments, thereby diminishing its optimality. Contrastingly, Path 3 emerges as the most rational and secure option. Such configuration effectively obviates aggressive path planning, enhances navigation safety, and still prioritizes the pursuit of a short path. Note that the USV is abstracted as a point whose volume can be incorporated by dilating the obstacles.}

\begin{figure*}[t]
\centering
\includegraphics[width=0.99\textwidth]{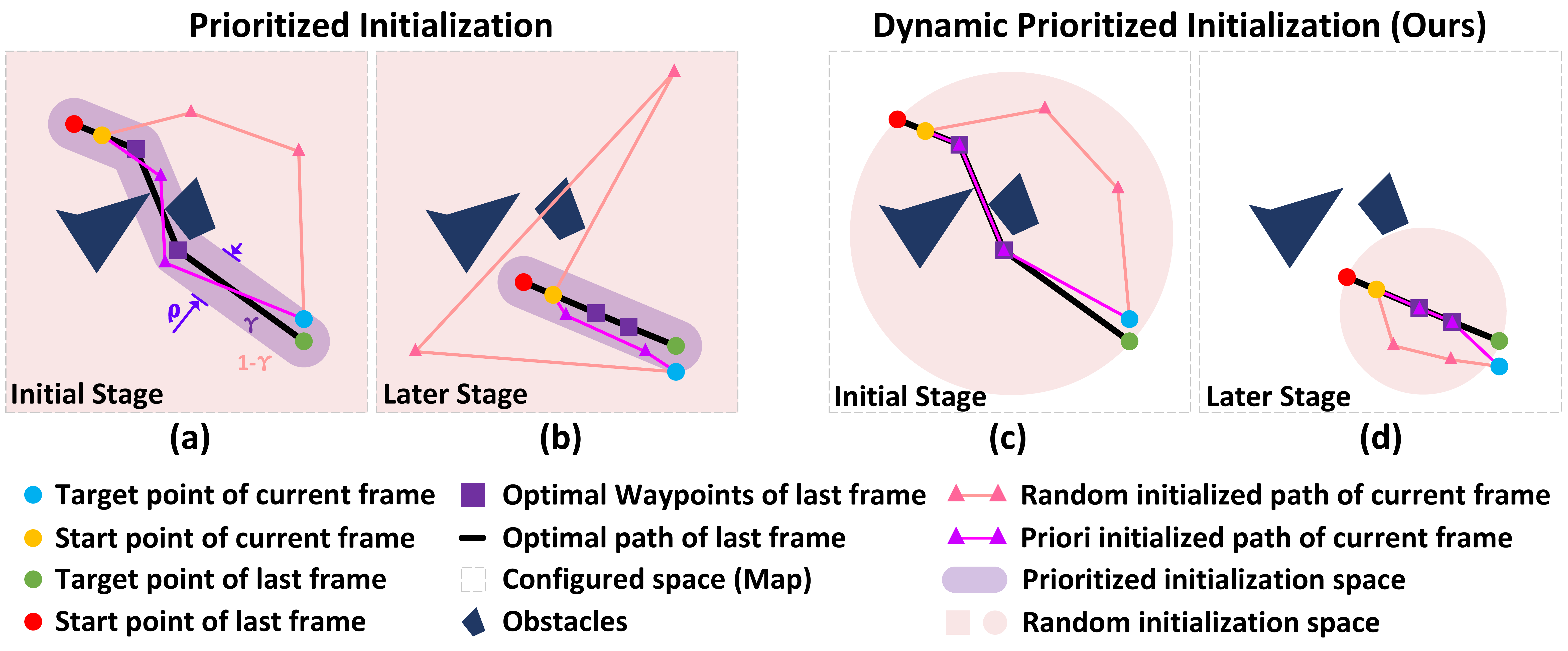}
\caption{Comparison between PI and DPI.}
\label{DPI}
\end{figure*}

In terms of the motion prediction for dynamic obstacles, various methodologies exist. A fancy way can be resorting to machine learning techniques, such as leveraging neural networks to forecast future trajectories \citep{predict1, predict2}. However, this approach may not be the most effective and practical due to the following reasons: 1) the computational demands of neural networks necessitate extra resources, posing a challenge for USVs with limited onboard computing capabilities; 2) the training of the prediction model requires additional efforts; 3) the generalization capability of the prediction model remains open to debate. In contrast, the advantages of our methods are evident: 1) the measurement of the obstacles' instantaneous velocities is tractable, generalizable, and computationally acceptable; 2) the kinematic segments can be seamlessly integrated into Eq. (\ref{obj}), with merely an extra term $\mu \cdot P(Z)^\nu$. 

Another issue that warrants consideration is the magnitude alignment between the length term $L(Z)$ and penalty terms. Neglecting such alignment can result in fluctuating performance at different navigation stages. An intuitive example is that the distance between the start and target point shrinks at the final navigation stage. This reduction renders the length term negligible in comparison to the penalty terms, thereby impairing the planning performance. A solution can be supplementing the penalty terms with a dynamic normalization factor, such that the penalty terms exhibit analogous decreasing characteristics relative to the length term. To this end, the objective function is normalized as follows:
\begin{equation}
\label{NDobj}
F(Z)=L(Z)+ \eta(\alpha \cdot Q(Z)^\beta + \mu \cdot P(Z)^\nu)
\end{equation}

\noindent where $\eta$ corresponds to the Euclidean distance between the current start and target point.

\subsection{Dynamic prioritized initialization}\label{Chaper:DPI}
As introduced in Section \ref{Chapter_PI}, the PI proposed by \cite{SEPSO} considerably ameliorates SEPSO's planning quality while expediting its convergence speed. Nevertheless, the PI is beset by two notable issues that detrimentally impact its performance in dynamic scenarios. Firstly, the PI fails to factor the shrinking nature of the navigation, leaving the random initialization space unmanaged throughout the entire navigation process. As revealed by Fig. \ref{DPI} (b), such configuration is unbefitting at the later stage of navigation, where the unmanaged random initialization space may engender noisy or even irrational initial paths that are less meaningful for the SEPSO to start on. Secondly, the PI has introduced two hyperparameters, $\gamma$ and $\rho$, whose determination relies on expert knowledge. Furthermore, these handcrafted hyperparameters also pose a potential threat to the robustness of the planning algorithm in dynamic environments. 

To address these issues, the DPI mechanism is proposed, as illustrated by Fig. \ref{DPI} (c) and (d). In particular, to accommodate the shrinking nature of the navigation, we narrow the random initialization space in accordance with the distance between the start and target points. More concretely, the unprioritized particles are randomly initialized within a circular area, with its diameter being the distance from the start point to the target point, as illuminated by Fig. \ref{DPI} (c) and (d). Concerning the fashion of prioritized initialization, our preliminary experiments indicate that inheriting the planned path from the last timestep for a portion of the population suffices to guide the whole swarm. In doing so, random initialization within the PI interval can be discarded, and the two hyperparameters $\gamma$ and $\rho$ can be omitted. In this context, at the onset of each planning timestep, each group of SEPSO employs one particle to inherit the previously planned path as the guidance for the current planning iteration. The remaining particles are then randomly initialized in the progressively narrowing circular area.

\begin{algorithm}[t]
\setstretch{1.2}
\SetAlgoNlRelativeSize{0}
\caption{Relaxation-based updating principle for the best values and positions}
\label{RS}
\KwIn{particle $X_{g,n}^k$}
\KwOut{the best positions $Pbest_{g,n}^k, Gbest_{g}^k, Tbest^k$ and the best values $U^{Pbest}_{g,n}, U^{Gbest}_{g}, U^{Tbest}$}
\tcp{$f(\cdot)$ is the fitness function; $R(a,b)$ denotes a uniform random variable between $(a,b)$}
\tcp{For minimization problems:}

\lIf{$f(X_{g,n}^k)< R(\lambda,1) \cdot U^{Pbest}_{g,n}$}{$Pbest_{g,n}^k \leftarrow X_{g,n}^k$} 
\lIf{$f(X_{g,n}^k)< U^{Pbest}_{g,n}$}{$U^{Pbest}_{g,n} \leftarrow f(X_{g,n}^k)$}

\lIf{$f(X_{g,n}^k)< R(\lambda,1) \cdot U^{Gbest}_{g}$}{$Gbest_{g}^k \leftarrow X_{g,n}^k$} 
\lIf{$f(X_{g,n}^k)< U^{Gbest}_{g}$}{$U^{Gbest}_{g} \leftarrow f(X_{g,n}^k)$}

\lIf{$f(X_{g,n}^k)< R(\lambda,1) \cdot U^{Tbest}$}{$Tbest^k \leftarrow X_{g,n}^k$} 
\lIf{$f(X_{g,n}^k)< U^{Tbest}$}{$U^{Tbest} \leftarrow f(X_{g,n}^k)$}
\end{algorithm}

\subsection{Autonomous hyperparameter tuning with relaxation strategy}\label{Chaper:RS}
Recalling Section \ref{SEPSO_preliminary}, during the evolution phase, the H-SEPSO is updated according to Algorithm \ref{best_principle}, with its fitness function given by Eq. (\ref{LFV}). However, Eq. (\ref{LFV}) is inapplicable for navigation tasks in dynamic environments due to its neglect of the collision cases and the environmental stochasticity. Since our pivotal objective is the determination of the search parameter that empowers safe and short navigations, a more comprehensive evaluation criterion for the planned path could be the total travel distance throughout the navigation. In this context, the determination of the search parameter can still be formulated as a minimization problem, with the fitness function of H-SEPSO defined as:
\begin{equation}
\label{score}
S(\theta) = 
\left\{
  \begin{array}{ll}
0 & \mbox{if Collide} \\
-1/\delta(\theta, \phi) & \mbox{if Arrive}
  \end{array}
\right.
\end{equation}

\noindent where $\delta$ is the total travel distance between fixed start and target points; $\phi$ is a random variable denoting the stochasticity of the environment. 

The presence of the random variable $\phi$ stems from the dependence of the total travel distance on both the capability of the optimization algorithm and the stochastic nature of the dynamic environment. That is, even the optimal search parameter, in certain instances, is likely to yield an unfavorable navigation trajectory due to the disturbances induced by dynamic obstacles. Similarly, a suboptimal search parameter could be overestimated when a relatively easy navigation task emerges. Evolving the search parameter $\theta$ with Eq. (\ref{score}) following Algorithm \ref{best_principle} can be problematic, as the overestimation could lead the whole population to exploit or even deadlock within a meaningless search space.

To mitigate the issue discussed above, a relaxation strategy, as given by Algorithm \ref{RS}, is proposed, wherein the update of the best positions $Pbest_{g,n}^k, Gbest_{g}^k, Tbest^k$ is subject to a relaxation strategy parameterized by $\lambda$. In the context of minimization, for problems with positive fitness value, $\lambda > 1$; whereas for problems with negative fitness value, $1 > \lambda >0$. This approach allows the update of $Pbest_{g,n}^k, Gbest_{g}^k, Tbest^k$ without necessitating a stringent decrease in fitness value, enabling a potentially advantageous particle to assist the population in escaping from the deadlock. It is imperative to highlight that the relaxation is not extended to the update of the best values $U^{Pbest}_{g,n}, U^{Gbest}_{g}, U^{Tbest}$, as doing so would impair the monotonic optimization characteristics the PSO-based techniques.

Finally, to facilitate autonomous tuning of the GPP-associated hyperparameters, ${[\alpha, \beta, \mu, \nu, \iota]}$ are integrated into Eq. (\ref{score}) as well. Correspondingly, the ultimate fitness function of H-SEPSO and the assembled hyperparameters are given as below:
\begin{equation}
\label{score_final}
S(\xi) = 
\left\{
  \begin{array}{ll}
0 & \mbox{if Collide} \\
-1/\delta(\xi, \phi) & \mbox{if Arrive}
  \end{array}
\right.
\end{equation}

\begin{equation}
\label{hyper}
\xi=[\alpha, \beta, \mu, \nu, \iota, \theta_1, \theta_2, \cdots, \theta_G]
\end{equation}

\subsection{\textcolor{black}{OkayPlan}}

\begin{figure}[t]
\centering
\includegraphics[width=0.4\textwidth]{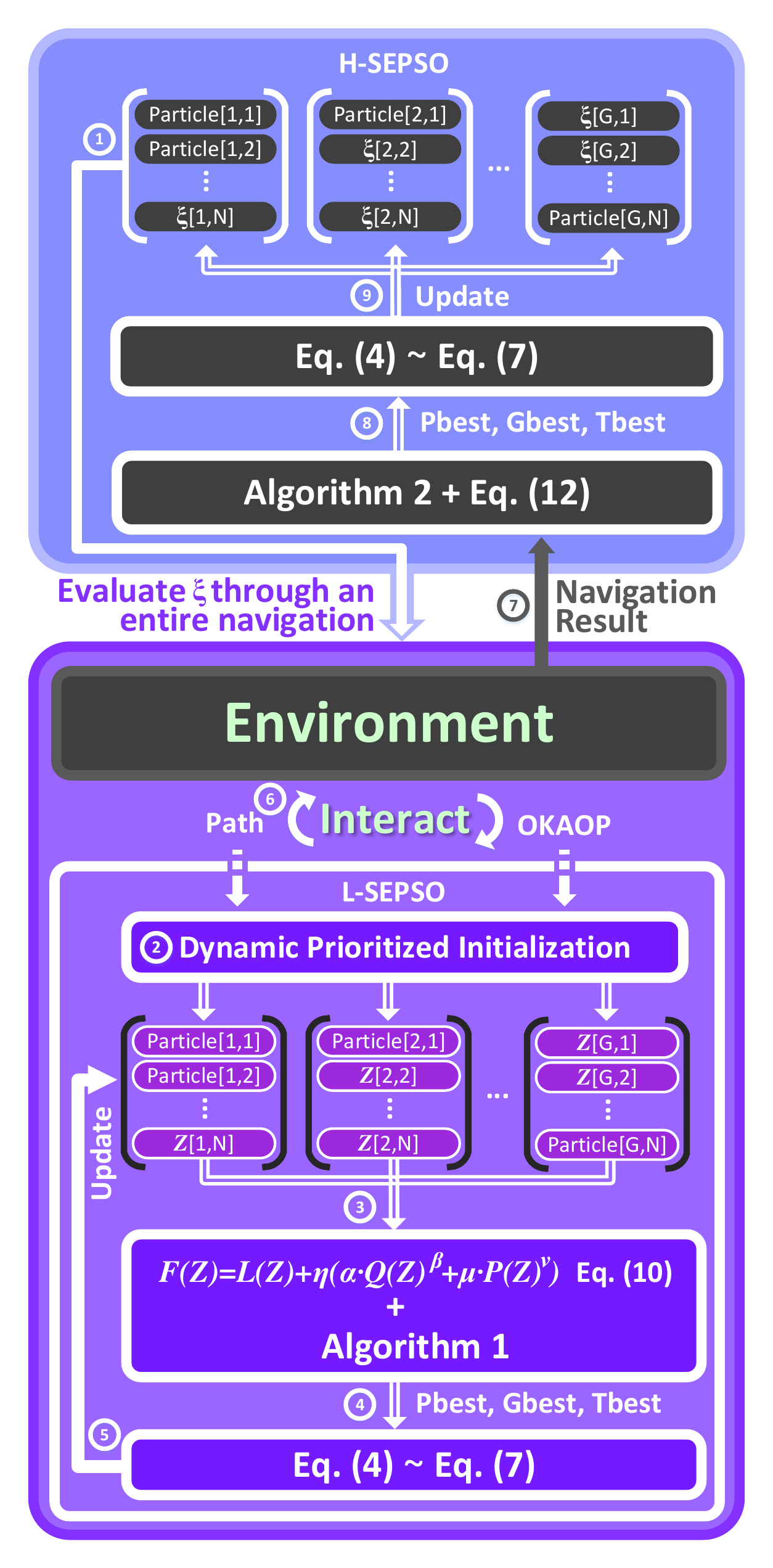}
\caption{\textcolor{black}{Diagram of the OkayPlan algorithm.}}
\label{dataflow}
\end{figure}

Assembling the components introduced in the preceding sections, we now formally present the OkayPlan algorithm, elucidated by Fig. \ref{dataflow}. Our algorithm encompasses two phases: the evolution phase and the application phase.

\subsubsection{\textcolor{black}{Evolution phase of OkayPlan}}\label{Evolution_phase}
The evolution phase is performed offline to tune OkayPlan's hyperparameters autonomously, whose pseudocode is given as follows: 

\noindent\ding{172} Pass one particle of H-SEPSO to be the hyperparameter of L-SEPSO.

\noindent\ding{173} Extract information from the environment and initialize the particles of L-SEPSO by DPI.

\noindent\ding{174} Evaluate the particles of L-SEPSO by Eq. (\ref{NDobj}).

\noindent\ding{175} Update the best positions and values of L-SEPSO via Algorithm \ref{best_principle}.

\noindent\ding{176} Update the particles of L-SEPSO via Eq. (\ref{DPPSO_V}) $\sim$ Eq. (\ref{params_theta}).

\noindent\ding{177} Iterate over \ding{174} $\sim$ \ding{176} for a maximum of 50 iterations or until a good path is identified. Then, control the USV to track the path before a new path is generated.

\noindent\ding{178} Repeat \ding{173} $\sim$ \ding{177} until the navigation is completed.

\noindent\ding{179} Execute \ding{172} $\sim$ \ding{178} for all the particles of H-SEPSO. Evaluate these navigation results by Eq. (\ref{score_final}). Update the best positions and values of H-SEPSO via Algorithm \ref{RS}.

\noindent\ding{180} Update the particles of H-SEPSO via Eq. (\ref{DPPSO_V}) $\sim$ Eq. (\ref{params_theta}).

\noindent\ding{181} Iterate over \ding{172} $\sim$ \ding{180} until H-SEPSO converges, and output its best particle as OkayPlan's hyperparameters for the application phase.

\subsubsection{\textcolor{black}{Application phase of OkayPlan}}
In the application phase, the H-SEPSO is discarded. The tuned hyperparameters from the evolution phase are now combined with the L-SEPSO for online global path planning. The pseudocode of the application phase is the same as \ding{173} $\sim$ \ding{177} delineated in the evolution phase.

\section{Experimental results}
In this section, we initially present two types of simulation environments. Utilizing these environments, we then demonstrate the hyperparameter tuning, evaluation, application, and ablation study of the OkayPlan. The specifics of the hardware platform employed to conduct these experiments are reported in Table \ref{hardware}.

\begin{table}[t]
\caption{Hardware platform for simulation experiments.}
\label{hardware}
  \begin{tabular*}{\tblwidth}{@{} LL@{}}
    \toprule
Component & Description\\
	\midrule
CPU   & Intel Core i9-13900KF  \\
GPU   & Nvidia RTX 4090 \\
RAM   & 64GB 5600MHz \\
System & Ubuntu 20.04.1 \\
	\bottomrule
  \end{tabular*}
\end{table}

\begin{figure*}[t]
\centering
\subfloat[Simple Case]{
    \includegraphics[width=0.185\textwidth]{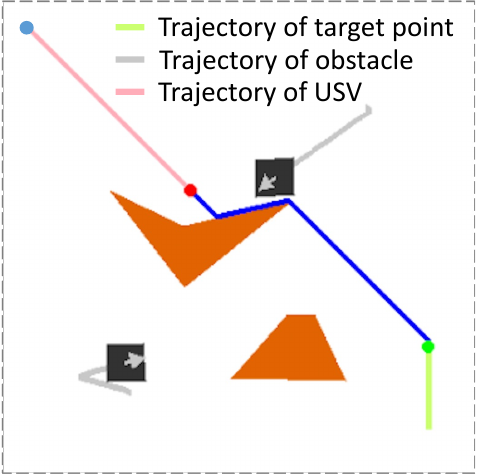}
    \label{CMO}}
\subfloat[Harbor]{
    \includegraphics[width=0.39\textwidth]{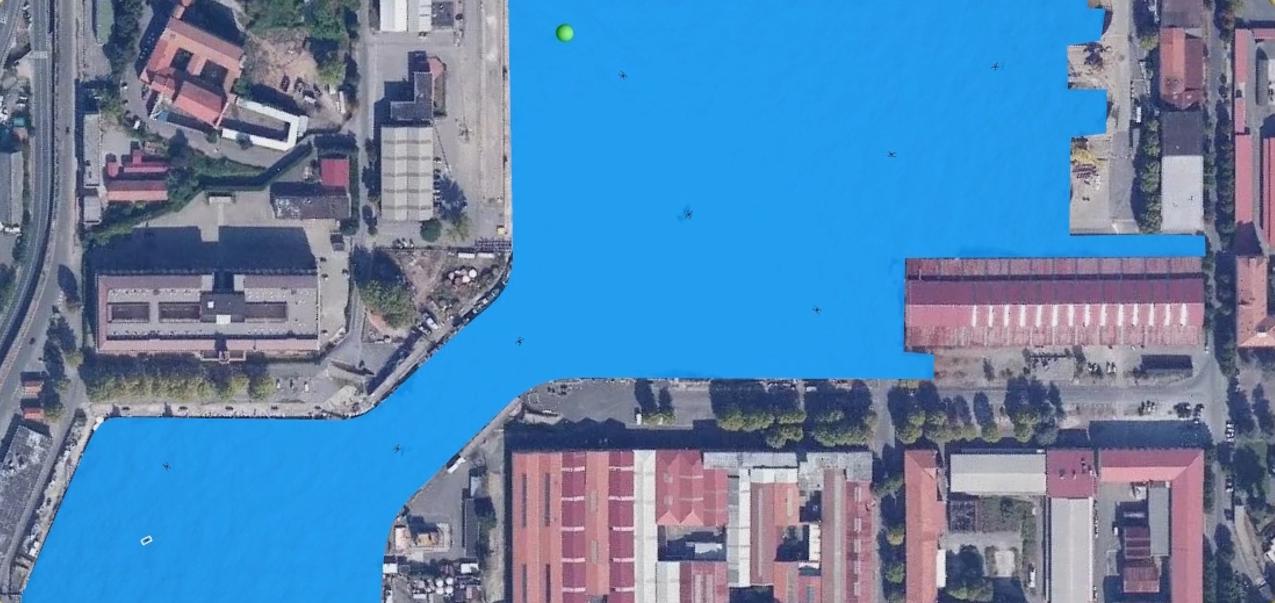}
    \label{Gazebo_harbor_A_1}}
\subfloat[USV]{
    \includegraphics[width=0.39\textwidth]{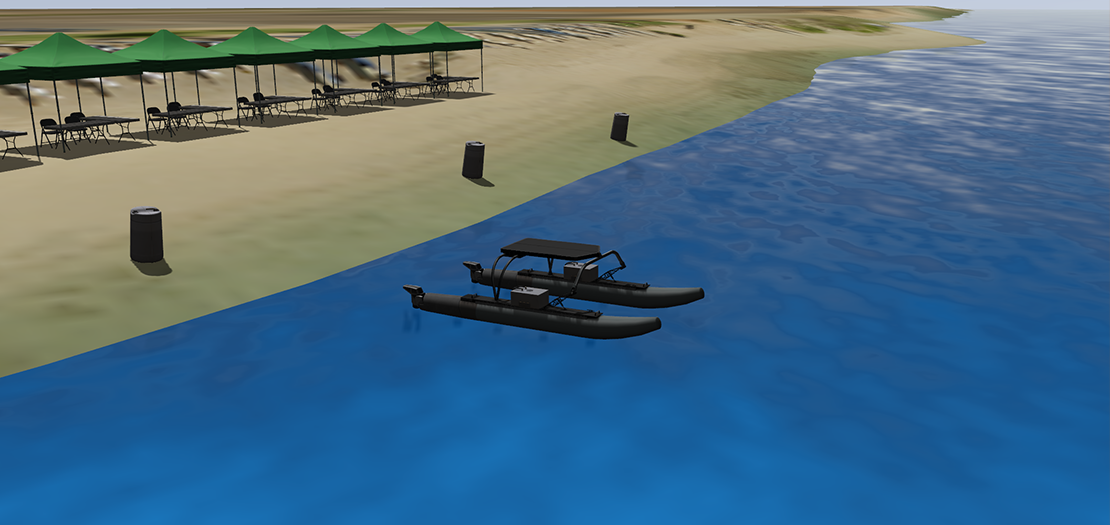}
    \label{USV_Outlook}}
\\
\subfloat[Complex Case]{
    \includegraphics[width=0.185\textwidth]{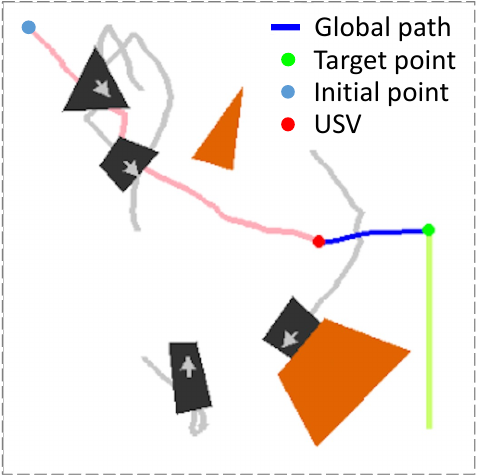}
    \label{RMO}}
\subfloat[Island]{
    \includegraphics[width=0.39\textwidth]{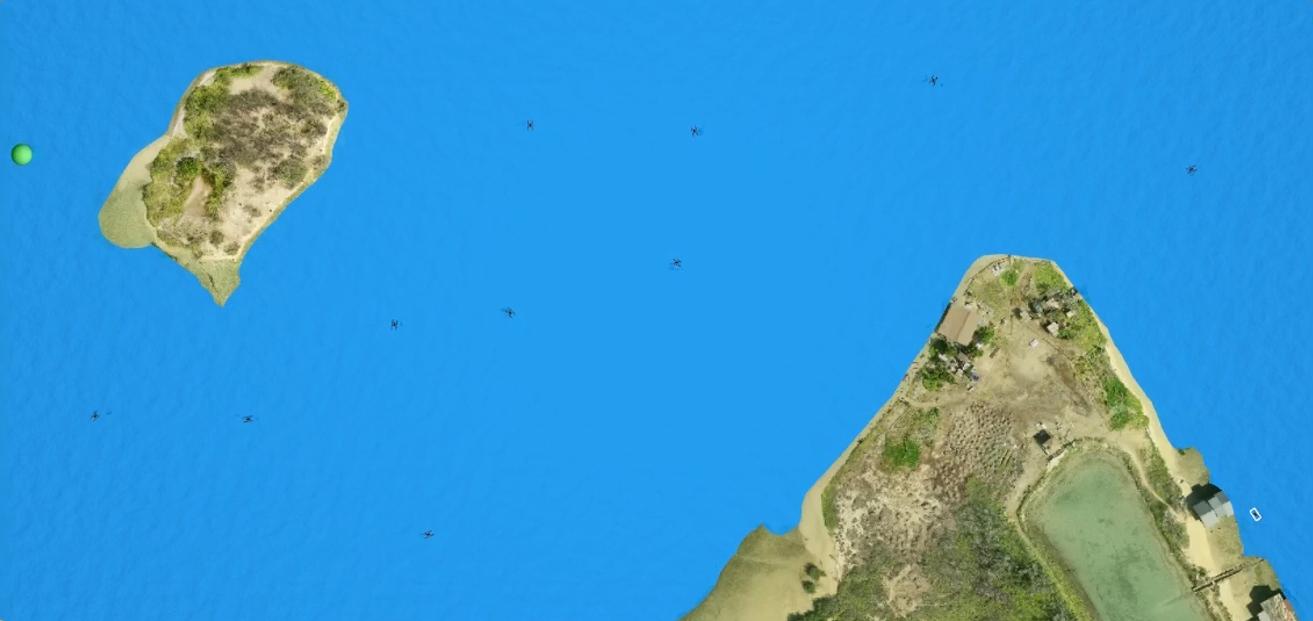}
    \label{Gazebo_island_B_1}}
\subfloat[The dimension of USV]{
    \includegraphics[width=0.39\textwidth]{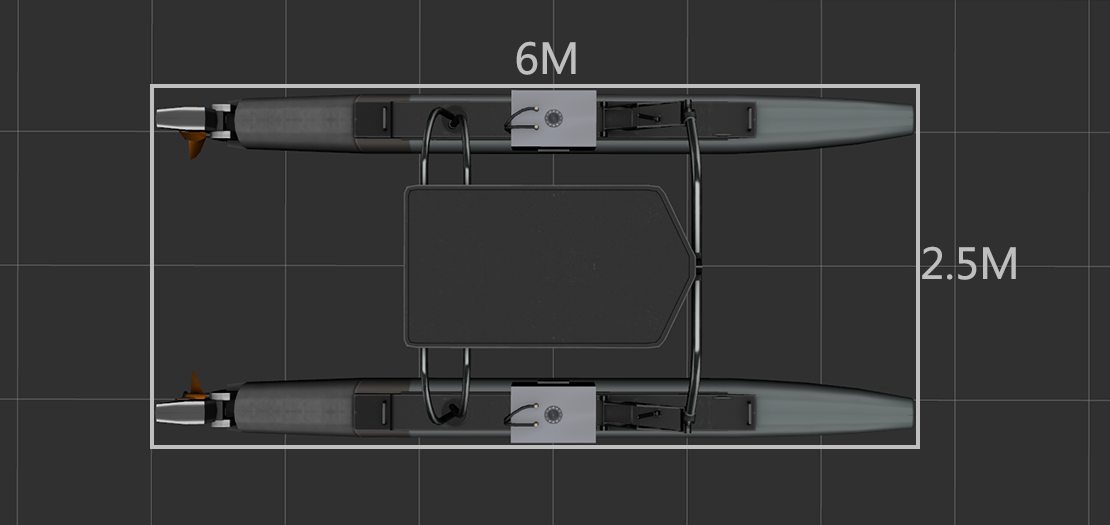}
    \label{USV_Scale}}
\caption{\textcolor{black}{Simulation environments. Block-based: (a) and (d). VRX: (b), (c), (e), and (f).}}
\label{env}
\end{figure*}

\begin{table*}[t]
\caption{\textcolor{black}{Basic information of the simulation environments.}}
\label{env_infos}
  \begin{tabular*}{\tblwidth}{@{} LLLLLLLLL@{}}
    \toprule
Environment & Width & Height & \makecell[l]{Static \\obstacles} & \makecell[l]{Dynamic\\obstacles} & \makecell[l]{Motion of \\ dynamic obstacles} & \makecell[l]{Obstacle \\velocity}& \makecell[l]{Controlled \\USV\ velocity } & \makecell[l]{Target\\ velocity} \\
	\midrule
Simple Case & $366\ m$ & $366\ m$ & Orange blocks & Black blocks & Consistently & $0\sim5.6 \ m/s$ & $6\ m/s$ & $3\ m/s$\\
Complex Case & $366\ m$ & $366\ m$ & Orange blocks & Black blocks & Randomly & $0\sim5.6 \ m/s$ & $6\ m/s$ & $3\ m/s$\\
Harbor & $720\ m$ & $344\ m$ & Harbor terrain& Other USVs & Randomly & $0\sim2.0 \ m/s$ & $8\ m/s$ & $3\ m/s$\\
Island & $620\ m$ & $296\ m$ & Island terrain& Other USVs & Randomly & $0\sim2.0 \ m/s$ & $8\ m/s$ & $1\ m/s$\\
	\bottomrule
  \end{tabular*}
\end{table*}

\begin{figure}[t]
\centering
\includegraphics[width=0.45\textwidth]{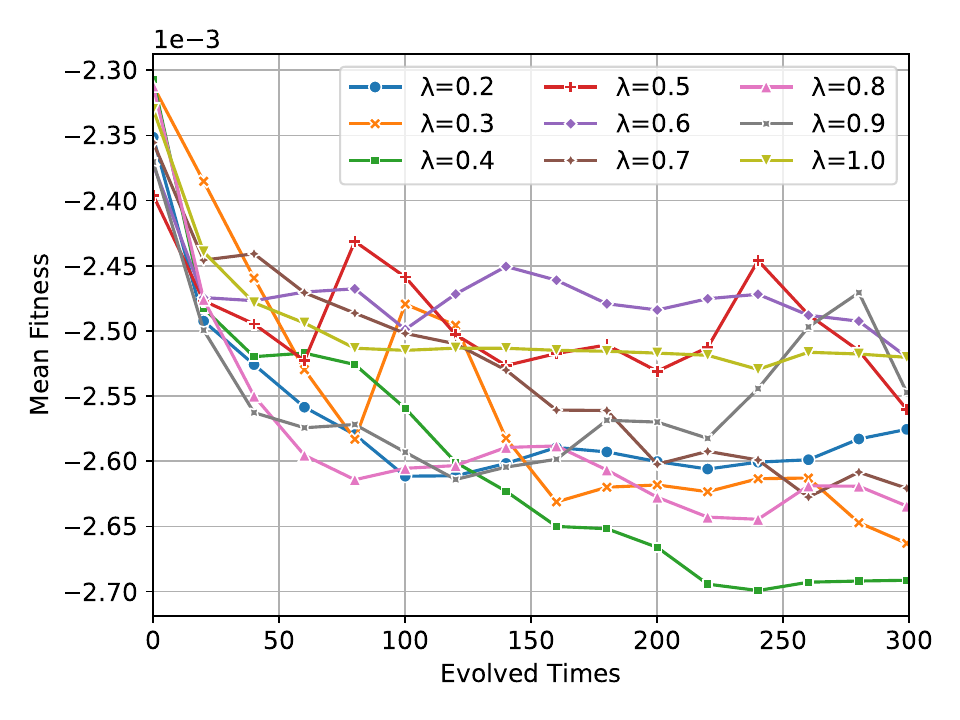}
\caption{Evolution curves of H-SEPSO with respect to different relaxation parameter. Here, every curve is averaged over 5 random seeds (0$\sim$4) to enhance credibility.}
\label{grid_search}
\end{figure}

\subsection{Simulation environments}
\textcolor{black}{To corroborate the superiority of OkayPlan, two distinct types of simulation environments were employed. Firstly, the Block-based environment \citep{SEPSO}, depicted in Fig. \ref{env} (a) and (d), serves as a fast and lightweight platform, facilitating the hyperparameter tuning of OkayPlan as well as the comparisons with other GPP algorithms. Secondly, the Virtual RobotX \citep[VRX;][]{VRX}, illustrated in Fig. \ref{env} (b), (c), (e), and (f), is a realistic simulation platform for USVs, which has been widely utilized in USV control and planning \citep{VRX_ctrl, VRX_plan}. The 3D physics simulations within VRX account for perturbations arising from environmental factors such as wind and waves, as well as the dynamic characteristics of USVs, including latency and drift, thereby providing a realistic testbed for assessing the performance of OkayPlan in the context of USV path planning. The details of these two simulation environments are presented in Table \ref{env_infos}.}

The task of these environments is to navigate the controlled USV from a fixed initial point to a moving target point. The controlled USV is abstracted as a point, while obstacles are represented by their respective bounding boxes. Noted that, in the VRX, the volume of the controlled USV is factored by dilating the obstacle bounding boxes by 6 $m$. Generally, such navigation tasks necessitate collaboration between GPP and LPP algorithms. However, given this paper's emphasis on the GPP algorithm, to avoid potential confounding effects from the LPP algorithm, we assume the controlled USV adheres strictly to the global path at a constant velocity of 6 $m/s$, thereby circumventing the need for LPP. Considering the dynamic nature of these environments, the GPP algorithms are executed consecutively at each timestep, replanning from the controlled USV's current position to the target point.

\subsection{Determination of hyperparameters}
\textcolor{black}{In this section, the relaxation strategy and  HSEF are utilized to determine OkayPlan's hyperparameters $\xi$, as the procedures discussed in Section \ref{Evolution_phase}.} We have adopted the experimental configurations from \cite{SEPSO}, wherein both L-SEPSO and H-SEPSO are partitioned into 8 groups to uphold population diversity. Each group of L-SEPSO is configured with 170 particles, with each particle represented as a 16-dimensional vector denoting the coordinates of 8 waypoints. In the case of H-SEPSO, each group is assigned with 10 particles. Each particle, namely the hyperparameter of L-SEPSO denoted by $\xi$, is a vector of length 53 (5 GPP parameters + 8 groups $\times$ 6 SEPSO parameters). Note that the search parameter of H-SEPSO employs the recommended configuration from the original paper \citep{SEPSO}.

Regarding the determination of the relaxation parameter $\lambda$, a coarse grid search from 0.2 to 1.0 with a resolution of 0.1 was conducted on the Complex Case. Within each search, the H-SEPSO was evolved for 300 times, and the mean fitness value over the entire population of H-SEPSO was recorded and drawn in Fig. \ref{grid_search}. The mean fitness value serves as a metric for the collective cognition of the particles in H-SEPSO. The premature convergence observed in the curve ($\lambda=1.0$) suggests that, without the relaxation strategy, the evolution of H-SEPSO is susceptible to entrapment in local optima. In contrast, the green curve ($\lambda=0.4$) distinctly illustrates the efficacy of the proposed relaxation strategy, indicating that a suitable relaxation parameter can substantially enhance the collective cognition of the population and facilitate the attainment of a superior solution.

The best hyperparameters identified by $\lambda=0.4$ are reported in Tables \ref{hyperSEPSO} and \ref{hyperGPP}, which will be employed to conduct the subsequent experiments. Here, the hyperparameters tuning for other environments have been omitted, as our forthcoming experiments will corroborate that these hyperparameters suffice to generalize across diverse cases.

\subsection{Evaluation of OkayPlan}
In this section, we seek to ascertain the advantages of the proposed OkayPlan in terms of path safety, length optimality, and computational efficiency versus existing GPP techniques. To this end, we have devised four statistical evaluation criteria to assess the navigation:

\begin{itemize}
 \item \textbf{Fitness}: Defined by Eq.(\ref{score_final}), where a lower fitness value signifies the GPP algorithm's capability to generate paths that are both shorter and safer.
 \item \textbf{Arrived Travel Distance}: The distance traveled when the USV successfully reaches the target. Note that the travel distance of the collision case is discarded. This metric serves as an indicator of the length optimality of the planned paths. 
 \item \textbf{Arrival Rate}: The rate at which the USV successfully navigates from the initial point to the target point, offering insights into the safety of the planned paths.
 \item \textbf{Time Per Planning}: The execution time of each planning averaged over the entire navigation, providing an assessment of the computational efficiency of the GPP algorithms evaluated.
\end{itemize}

To facilitate a comprehensive comparison, we have prepared 9 contemporary and canonical GPP algorithms, spanning metaheuristic-based, sample-based, and search-based approaches. The particulars of these algorithms are exhibited in Table \ref{GPPalgorithms}. Harnessing the hyperparameters specified in Tables \ref{hyperSEPSO} and \ref{hyperGPP}, the proposed OkayPlan is compared against these 9 algorithms within the Simple Case and the Complex Case. Snapshots illustrating their respective navigation processes are exhibited in Fig. \ref{CMO_traj} and Fig. \ref{RMO_traj} within the Appendix. For a rigorous analysis, these experiments are repeated 100 times, with random seeds ranging from 0 to 99. The averaged results are then presented in Tables \ref{SimpleCaseResult} and \ref{ComplexCaseResult}. 

\subsubsection{\textcolor{black}{Path safety}}
The \textit{Arrival Rate} presented in Tables \ref{SimpleCaseResult} and \ref{ComplexCaseResult} demonstrate the superior path safety of OkayPlan, as it successfully navigated the controlled USV to the target point in all 100 trials, while other benchmark algorithms experienced failures, particularly in the Complex Case. This enhanced safety can be attributed to the proposed OKAOP, wherein future obstacle motions are explicitly taken into account during planning, resulting in a unique conservative planning behavior. This conclusion can be further substantiated by \textit{timestep 15} in Fig. \ref{CMO_traj}, where OkayPlan avoids potential conflicts with dynamic obstacles by circumventing them. In contrast, the other GPP algorithms generate aggressive paths, thereby increasing collision risks and compromising overall navigation safety.

\subsubsection{\textcolor{black}{Path length optimality}}
\begin{figure*}[h]
\centering
\includegraphics[width=0.88\textwidth]{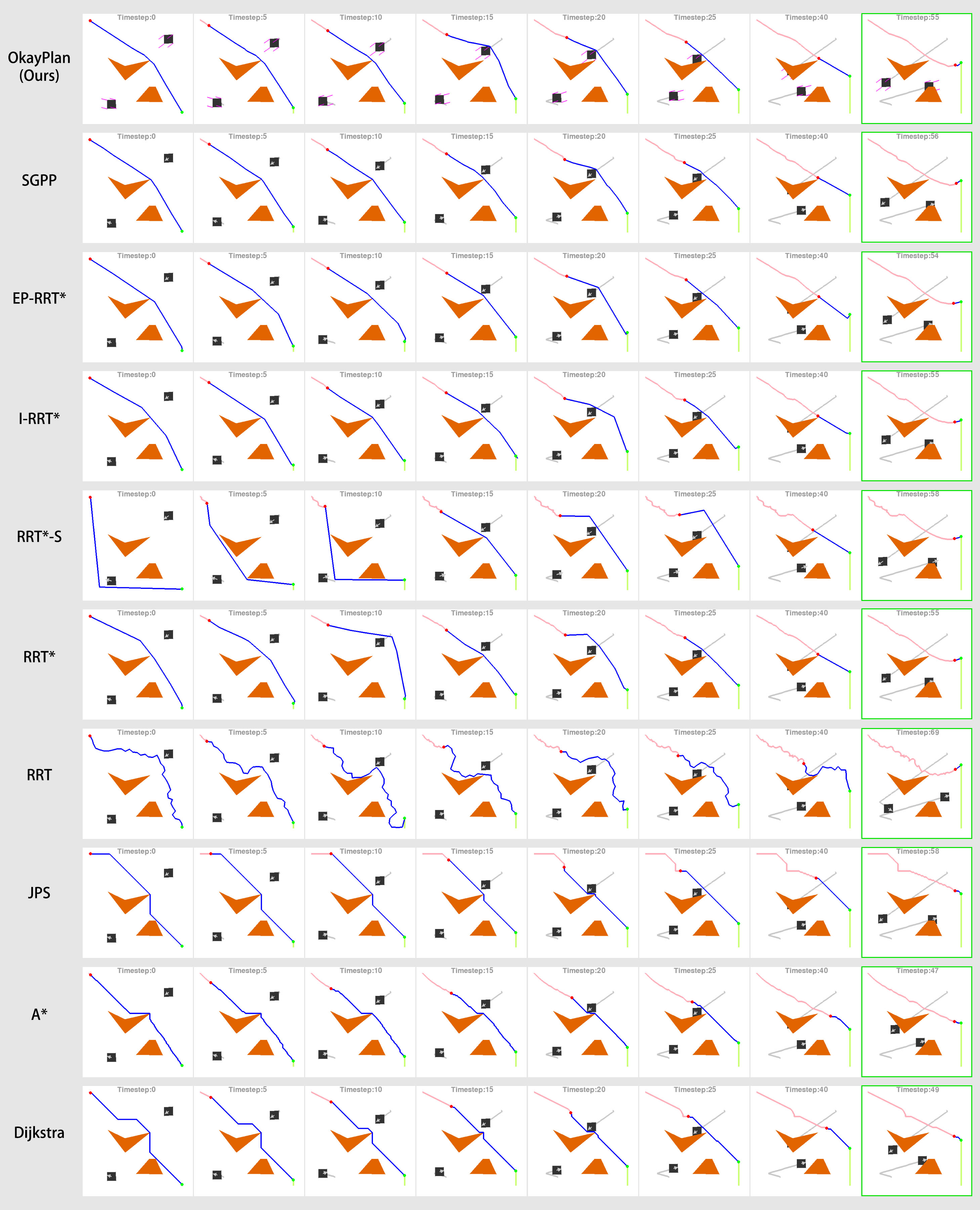}
\caption{\textcolor{black}{Behaviors comparison between different GPP algorithms on \textbf{Simple Case}. Each row represents the planning results of a specific GPP algorithm at different timesteps. The temporal sequence progresses from left to right. On the rightmost side, the algorithm that successfully reaches the target point is marked with a green box. OkayPlan's unique conservative planning strategy to avoid contention with the obstacles can be observed at timestep 15, where other algorithms choose aggressive planning, risking their path safety.}}
\label{CMO_traj}
\end{figure*}

\begin{figure*}[h]
\centering
\includegraphics[width=0.88\textwidth]{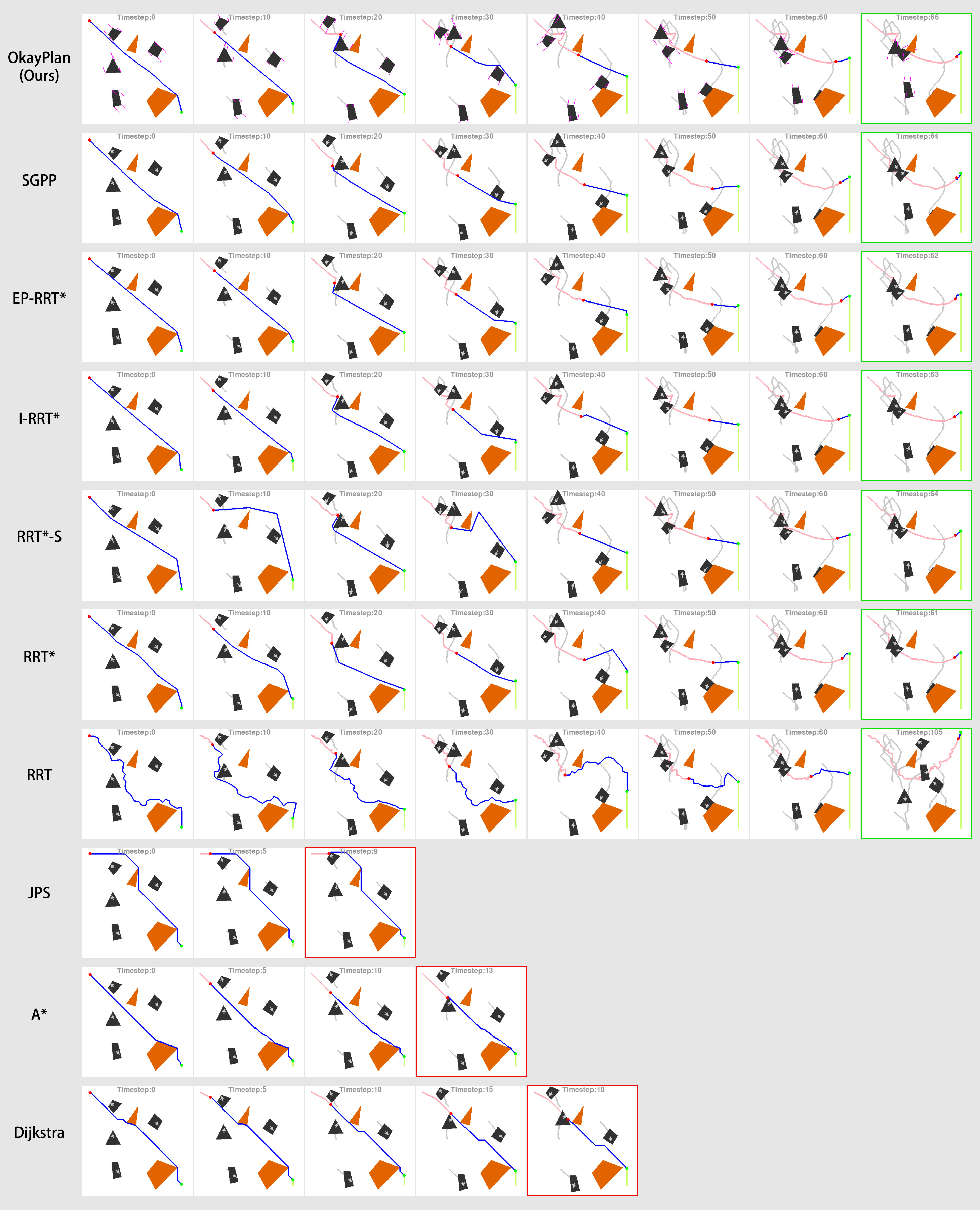}
\caption{\textcolor{black}{Behaviors comparison between different GPP algorithms on \textbf{Complex Case}. Each row of snapshots represents the planning results of a specific GPP algorithm at different timesteps. The temporal sequence progresses from left to right. On the rightmost side, the algorithm that successfully reaches the target point is marked with a green box, while algorithms that experience collisions are marked with a red box.}}
\label{RMO_traj}
\end{figure*}

The \textit{Fitness} reported in Tables \ref{SimpleCaseResult} and \ref{ComplexCaseResult}, along with the path trajectories in Fig. \ref{CMO_traj} and Fig. \ref{RMO_traj}, highlight the commendable path length optimality achieved by the proposed OkayPlan in dynamic environments. However, a marginal performance decline in \textit{Arrived Travel Distance} can also be noticed. We attribute this decline to OkayPlan's unique conservative planning. As exemplified by Paths 1 and 3 in Fig. \ref{OKA}, a safe path may compromise the length optimality. We argue that this trade-off is acceptable and necessary, as it effectively prevents the USV from colliding with dynamic obstacles and prominently elevates the success rate of navigation.

\begin{table}[t]
\caption{SEPSO-associated hyperparameters.}
\label{hyperSEPSO}
\resizebox{0.4\textwidth}{!}{
  \begin{tabular*}{\tblwidth}{@{} LLLLLLL@{}}
    \toprule
    Group & $\omega_{init}$ & $\omega_{end}$ & $V_{limit}$ & $C1$    & $C2$    & $C3$\\
	\midrule
    1     & 0.9000 & 0.9000 & 0.1000 & 1.0000 & 2.0000 & 1.0000 \\
    2     & 0.2000 & 0.1000 & 0.1000 & 1.4853 & 1.0000 & 1.0000 \\
    3     & 0.7434 & 0.9000 & 0.1389 & 1.0000 & 1.0000 & 2.0000 \\
    4     & 0.9000 & 0.9000 & 0.1000 & 1.0756 & 1.0000 & 1.2968 \\
    5     & 0.2000 & 0.9000 & 0.8000 & 2.0000 & 2.0000 & 2.0000 \\
    6     & 0.6094 & 0.1000 & 0.1000 & 1.0000 & 1.3316 & 2.0000 \\
    7     & 0.8271 & 0.1000 & 0.8000 & 2.0000 & 2.0000 & 1.0000 \\
    8     & 0.9000 & 0.7743 & 0.8000 & 1.9968 & 1.9253 & 1.0000 \\
	\bottomrule
  \end{tabular*}}
\end{table}

\begin{table}[t]
\caption{GPP-associated hyperparameters.}
\label{hyperGPP}
\resizebox{0.4\textwidth}{!}{
  \begin{tabular*}{\tblwidth}{@{} LLLLL@{}}
    \toprule
$\alpha$ & $\beta$ & $\mu$ & $\nu$ & $\iota$\\
	\midrule
    4.0000& 1.0000& 3.9827& 6.0000& 5.2032  \\
	\bottomrule
  \end{tabular*}}
\end{table}

\begin{table*}[t]
\caption{\textcolor{black}{Details of global path planning algorithms.}}
\label{GPPalgorithms}
\resizebox{0.95\textwidth}{!}{
  \begin{tabular*}{\tblwidth}{@{} LLLL@{}}
    \toprule
    Algorithm & Abbreviation & Type  & Reference \\
	\midrule
    OkayPlan & OkayPlan & Metaheuristic-based & This paper \\
    SEPSO-based Global Path Planning & SGPP & Metaheuristic-based & \citet{SEPSO} \\
    Expanding Path RRT* & EP-RRT*  & Sample-based & \citet{EP_RRTstar} \\
    Informed RRT* & I-RRT*  & Sample-based & \citet{IRRTstar} \\
    RRT*-Smart & RRT*-S  & Sample-based & \citet{RRTstar_smart} \\
    RRT* & RRT*  & Sample-based & \citet{RRTstar} \\
    Rapidly-exploring random trees & RRT   & Sample-based & \citet{RRT} \\
    Jump Point Search & JPS    & Search-based & \citet{JPS} \\
    A* & A*    & Search-based & \citet{Astar} \\
    Dijkstra & Dijkstra & Search-based & \citet{Dijkstra} \\
	\bottomrule
  \end{tabular*}}
\end{table*}

\begin{table*}[t]
\caption{\textcolor{black}{Comparison on \textbf{Simple Case}.}}
\label{SimpleCaseResult}
\resizebox{0.95\textwidth}{!}{
  \begin{tabular*}{\tblwidth}{@{} LLLLLL@{}}
    \toprule
    Algorithm & Metric & Fitness ($\times 10^{-3}$) & Arrived Travel Distance $(m)$   & Time Per Planning ($s$)  & Arrival Rate \\
	\midrule
    \multirow{2}[0]{*}{OkayPlan} & mean  & \textbf{-2.79 } & 366.18  & \textbf{0.008 } & \multirow{2}[0]{*}{\textbf{100\% }} \\
          & std.  & 0.36  & 62.21  & 0.001  &  \\
    \multirow{2}[0]{*}{SGPP} & mean  & -2.24  & \textbf{351.08 } & 0.009  & \multirow{2}[0]{*}{78\% } \\
          & std.  & 1.20  & 33.95  & 0.001  &  \\
    \multirow{2}[0]{*}{EP-RRT*} & mean  & -2.21  & 360.98  & 0.014  & \multirow{2}[0]{*}{78\% } \\
          & std.  & 1.13  & 5.51  & 0.016  &  \\
    \multirow{2}[0]{*}{I-RRT*} & mean  & -2.13  & 361.77  & 0.019  & \multirow{2}[0]{*}{77\% } \\
          & std.  & 1.02  & 6.12  & 0.013  &  \\
    \multirow{2}[0]{*}{RRT*-S} & mean  & -2.67  & 360.50  & 0.452  & \multirow{2}[0]{*}{96\% } \\
          & std.  & 0.55  & 13.87  & 0.024  &  \\
    \multirow{2}[0]{*}{RRT*} & mean  & -2.08  & 365.37  & 0.023  & \multirow{2}[0]{*}{76\% } \\
          & std.  & 1.17  & 5.47  & 0.012  &  \\
    \multirow{2}[0]{*}{RRT} & mean  & -1.69  & 495.04  & 0.010  & \multirow{2}[0]{*}{83\% } \\
          & std.  & 0.78  & 39.51  & 0.002  &  \\
    \multirow{2}[0]{*}{JPS} & mean  & -2.53  & 355.80  & 0.137  & \multirow{2}[0]{*}{90\% } \\
          & std.  & 0.84  & 6.54  & 0.007  &  \\
    \multirow{2}[0]{*}{A*} & mean  & -2.04  & 381.54  & 1.174  & \multirow{2}[0]{*}{78\% } \\
          & std.  & 1.09  & 4.20  & 0.262  &  \\
    \multirow{2}[0]{*}{Dijkstra} & mean  & -1.95  & 375.13  & 3.511  & \multirow{2}[0]{*}{73\% } \\
          & std.  & 1.18  & 3.59  & 0.488  &  \\
	\bottomrule
  \end{tabular*}}
\end{table*}

\begin{table*}[t]
\caption{\textcolor{black}{Comparison on \textbf{Complex Case}.}}
\label{ComplexCaseResult}
\resizebox{0.95\textwidth}{!}{
  \begin{tabular*}{\tblwidth}{@{} LLLLLL@{}}
    \toprule
    Algorithm & Metric & Fitness ($\times 10^{-3}$) & Arrived Travel Distance $(m)$   & Time Per Planning ($s$)  & Arrival Rate \\
	\midrule
    \multirow{2}[0]{*}{OkayPlan} & mean  & \textbf{-2.75 } & 365.10 & 0.009  & \multirow{2}[0]{*}{\textbf{100\% }} \\
          & std.  & 0.14  & 21.56  & 0.001  &  \\
    \multirow{2}[0]{*}{SGPP} & mean  & -1.35  & 391.73  & \textbf{0.008 } & \multirow{2}[0]{*}{52\% } \\
          & std.  & 1.31  & 60.11  & 0.002  &  \\
    \multirow{2}[0]{*}{EP-RRT*} & mean  & -1.29  & 367.13  & 0.027  & \multirow{2}[0]{*}{51\% } \\
          & std.  & 1.29  & 8.91  & 0.023  &  \\
    \multirow{2}[0]{*}{I-RRT*} & mean  & -1.21  & 368.21  & 0.033  & \multirow{2}[0]{*}{46\% } \\
          & std.  & 1.34  & 10.21  & 0.018  &  \\
    \multirow{2}[0]{*}{RRT*-S} & mean  & -1.28  & 388.41  & 0.378  & \multirow{2}[0]{*}{49\% } \\
          & std.  & 1.32  & 58.13  & 0.078  &  \\
    \multirow{2}[0]{*}{RRT*} & mean  & -1.19  & 369.00  & 0.051  & \multirow{2}[0]{*}{44\% } \\
          & std.  & 1.35  & 9.09  & 0.014  &  \\
    \multirow{2}[0]{*}{RRT} & mean  & -0.93  & 518.38  & 0.018  & \multirow{2}[0]{*}{48\% } \\
          & std.  & 0.97  & 41.00  & 0.005  &  \\
    \multirow{2}[0]{*}{JPS} & mean  & -1.60  & \textbf{356.32}  & 0.117  & \multirow{2}[0]{*}{57\% } \\
          & std.  & 1.39  & 5.48  & 0.022  &  \\
    \multirow{2}[0]{*}{A*} & mean  & -1.35  & 379.12  & 1.609  & \multirow{2}[0]{*}{51\% } \\
          & std.  & 1.32  & 3.28  & 0.696  &  \\
    \multirow{2}[0]{*}{Dijkstra} & mean  & -1.20  & 375.82  & 4.111  & \multirow{2}[0]{*}{45\% } \\
          & std.  & 1.32  & 4.08  & 0.567  &  \\
	\bottomrule
  \end{tabular*}}
\end{table*}

\begin{figure}[t]
\centering
\includegraphics[width=0.4\textwidth]{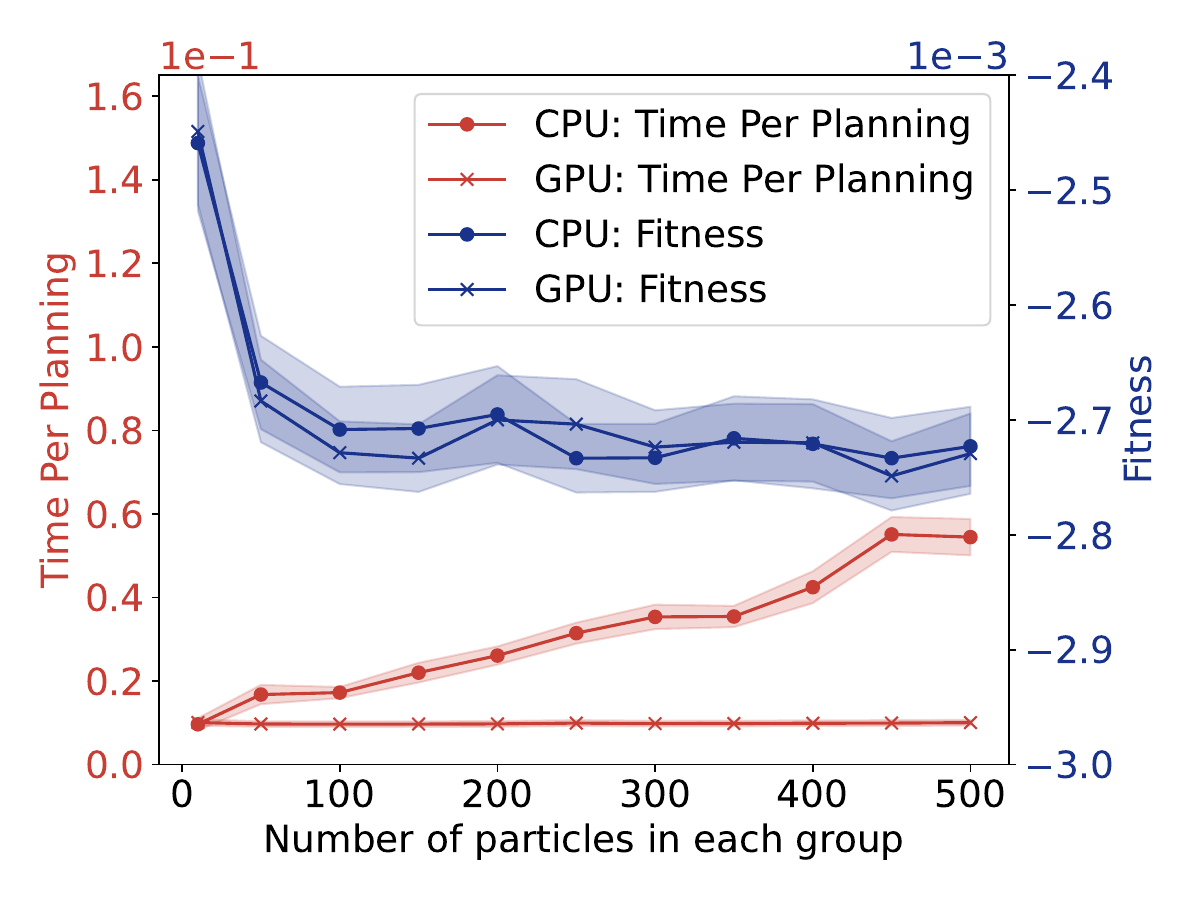}
\caption{\textcolor{black}{OkayPlan's performance under different conditions.}}
\label{NaP}
\end{figure}

\subsubsection{\textcolor{black}{Computational efficiency}}

In both Simple and Complex Cases, OkayPlan exhibits competitive computational efficiency compared to its predecessor, SGPP. This observation suggests that the proposed OKAOP and DPI have been seamlessly integrated without imposing additional computational burdens. Impressively, OkayPlan attains a high planning frequency of 125 Hz (0.008 \textit{s/p}), validating its capability to efficiently address real-time planning tasks in dynamic environments.

\textcolor{black}{It should be noted that the performance of OkayPlan can be influenced by the number of particles employed as well as the computational platform utilized. To provide insights into this, Fig. \ref{NaP} presents an analysis of OkayPlan's performance under varying conditions. The results are averaged over 1000 trials, with the solid curves representing the mean values and the translucent areas depicting the 95\% confidence intervals. In general, a larger number of particles tends to yield better planning results. Notably, OkayPlan's computation time remains stable on a GPU as the particle number mounts. Although the computing speed decreases with an increasing number of particles when running on a CPU, a laudable planning result can still be obtained at a planning frequency of approximately 50 Hz (0.02 \textit{s/p}) by employing 100 particles per group. This finding suggests that OkayPlan is also well-suited for USVs with limited onboard computational resources without GPUs.}

\begin{table*}[h]
\caption{Ablation Study on \textbf{Simple Case}.}
\label{SimpleAblation}
  \begin{tabular*}{\tblwidth}{@{} LLLLLL@{}}
    \toprule
    Algorithm & Metric & Fitness ($\times 10^{-3}$) & Arrived Travel Distance $(m)$   & Time Per Planning ($s$)  & Arrival Rate \\
	\midrule
    \multirow{2}[0]{*}{OkayPath} & mean  & -2.79  & 366.18  & 0.008  & \multirow{2}[0]{*}{100\% } \\
          & std.  & 0.36  & 62.21  & 0.001  &  \\
    \multirow{2}[0]{*}{No OKAOP} & mean  & -2.35  & 358.55  & 0.009  & \multirow{2}[0]{*}{83\% } \\
          & std.  & 1.10  & 51.42  & 0.002  &  \\
    \multirow{2}[0]{*}{No Dynamic Normalization} & mean  & -2.78  & 369.54  & 0.009  & \multirow{2}[0]{*}{100\% } \\
          & std.  & 0.37  & 70.31  & 0.001  &  \\
    \multirow{2}[0]{*}{No DPI} & mean  & -2.72  & 371.58  & 0.010  & \multirow{2}[0]{*}{100\% } \\
          & std.  & 0.24  & 40.92  & 0.001  &  \\
    \multirow{2}[0]{*}{No Relaxation Strategy} & mean  & -2.77  & 370.68  & 0.009  & \multirow{2}[0]{*}{100\% } \\
          & std.  & 0.37  & 74.11  & 0.001  &  \\
	\bottomrule
  \end{tabular*}
\end{table*}

\begin{table*}[h]
\caption{Ablation Study on \textbf{Complex Case}.}
\label{ComplexAblation}
  \begin{tabular*}{\tblwidth}{@{} LLLLLL@{}}
    \toprule
    Algorithm & Metric & Fitness ($\times 10^{-3}$) & Arrived Travel Distance $(m)$   & Time Per Planning ($s$)  & Arrival Rate \\
	\midrule
    \multirow{2}[0]{*}{OkayPath} & mean  & -2.75  & 365.10  & 0.009  & \multirow{2}[0]{*}{100\% } \\
          & std.  & 0.14  & 21.56  & 0.001  &  \\
    \multirow{2}[0]{*}{No OKAOP} & mean  & -1.58  & 349.09  & 0.010  & \multirow{2}[0]{*}{55\% } \\
          & std.  & 1.43  & 6.78  & 0.002  &  \\
    \multirow{2}[0]{*}{No Dynamic Normalization} & mean  & -2.71  & 367.70  & 0.010  & \multirow{2}[0]{*}{99\% } \\
          & std.  & 0.32  & 26.66  & 0.001  &  \\
    \multirow{2}[0]{*}{No DPI} & mean  & -2.18  & 464.72  & 0.010  & \multirow{2}[0]{*}{97\% } \\
          & std.  & 0.57  & 105.53  & 0.001  &  \\
    \multirow{2}[0]{*}{No Relaxation Strategy} & mean  & -2.59  & 387.82  & 0.009  & \multirow{2}[0]{*}{99\% } \\
          & std.  & 0.39  & 56.15  & 0.001  &  \\
	\bottomrule
  \end{tabular*}
\end{table*} 

\begin{figure*}[t]
\centering
\includegraphics[width=1\textwidth]{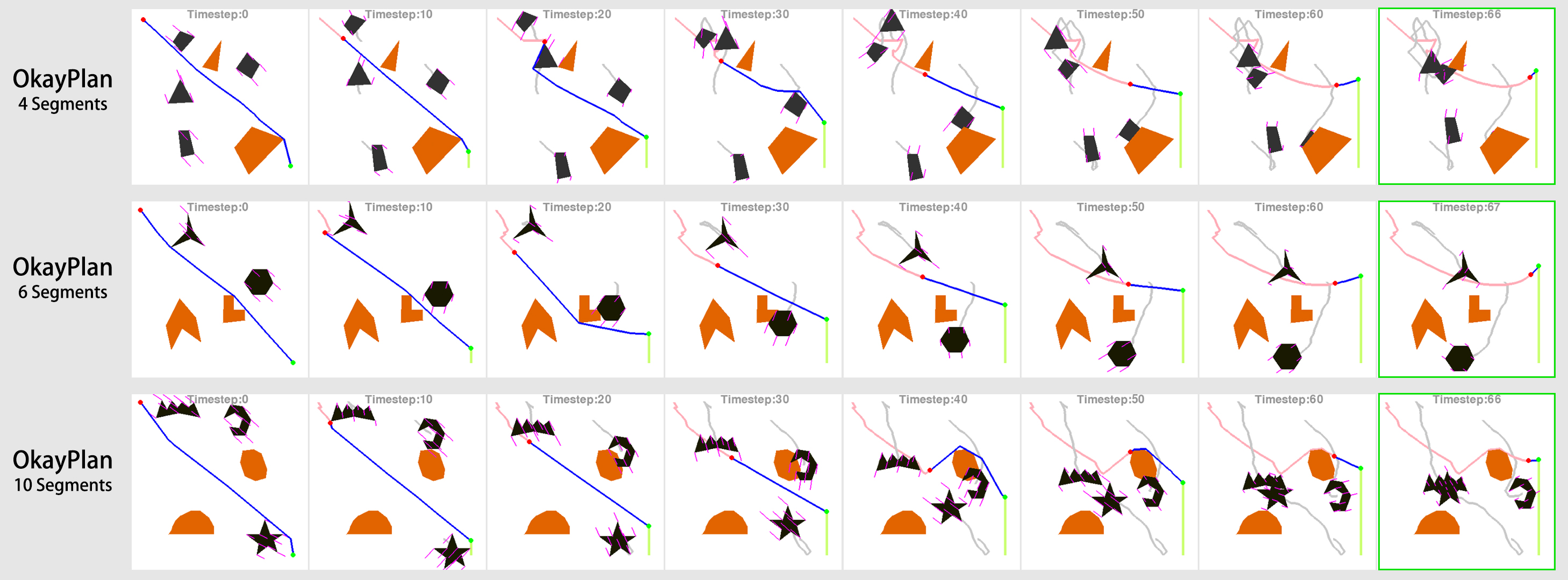}
\caption{\textcolor{black}{Experiments with intricate obstacle shapes.}}
\label{Segment}
\end{figure*}

\subsubsection{\textcolor{black}{Obstacle shape}}
\textcolor{black}{In previous sections, the obstacles are represented by bounding boxes with 4 segments. However, in the context of marine environments, such settings may give rise to fidelity losses when dealing with complex terrains. Thus, we have supplemented our experiments with bounding boxes comprising more segments to demonstrate OkayPlan's compatibility with intricate obstacle shapes, as the results show in Fig. \ref{Segment}.}


\begin{figure*}[t]
\centering
\includegraphics[width=1\textwidth]{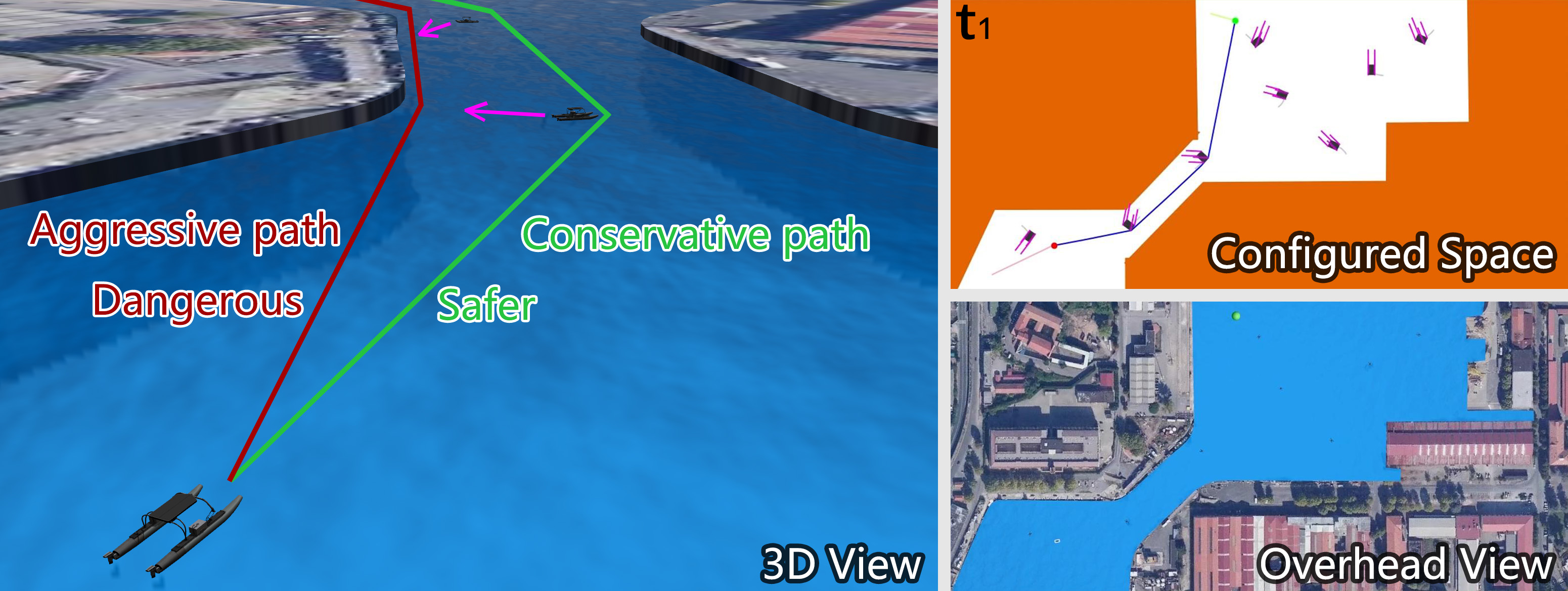}
\caption{\textcolor{black}{OkayPlan's planning result (green) on harbor environment.}}
\label{3Dshow}
\end{figure*}

\subsection{Ablation studies}
In this section, we conduct ablation studies on OkayPlan by removing each of its key components to explicate their individual contributions. The experimental setups for these ablation studies are as follows:

\begin{itemize}
\item \textbf{No OKAOP} (Section \ref{Chaper:OKA}): $\mu$ in Eq. (\ref{NDobj}) is set to be 0.
\item \textbf{No Dynamic Normalization} (Section \ref{Chaper:OKA}): $\eta$ in Eq. (\ref{NDobj}) is set to be the Euclidean distance between the initial point and target point at the beginning of the navigation and remains constant thoughout the navigation.
\item  \textbf{No DPI} (Section \ref{Chaper:DPI}): the DPI mechanism is degenerated into the PI mechanism \citep{SEPSO} as illustrated in Fig. \ref{DPI} (a) and (b).
\item  \textbf{No Relaxation Strategy} (Section \ref{Chaper:RS}): OkayPlan employs the hyperparameter $\xi$ evolved with relaxation parameter $\lambda=1$.
\end{itemize}

The ablation studies are performed on both the Simple Case and Complex Case. Each of them is repeated 100 times with random seeds ranging from 1 to 99, and the averaged results are reported in Tables \ref{SimpleAblation} and \ref{ComplexAblation}. The results suggest that the four components under investigation exhibit no explicit impact on computational efficiency, mirroring the compact and integral design of OkayPlan. Omitting the OKAOP results in a significant decrease in the \textit{Arrival Rate}, revealing its remarkable contribution to the safety of the planned paths. Meanwhile, as discussed in the preceding section, a trade-off between path safety and length optimality is observed when ablating the OKAOP, which validates our previous conjecture. While the Dynamic Normalization contributes marginally to OkayPlan, its incorporation is still considered beneficial due to its negligible effect on the computational overhead. The DPI and the Relaxation Strategy manifest similar repercussions, where their exclusion leads to a loss of length optimality, as indicated by the \textit{Fitness} and \textit{Arrived Travel Distance}.


\subsection{\textcolor{black}{Application to marine navigation}}
\begin{figure*}[h]
\centering
\includegraphics[width=0.88\textwidth]{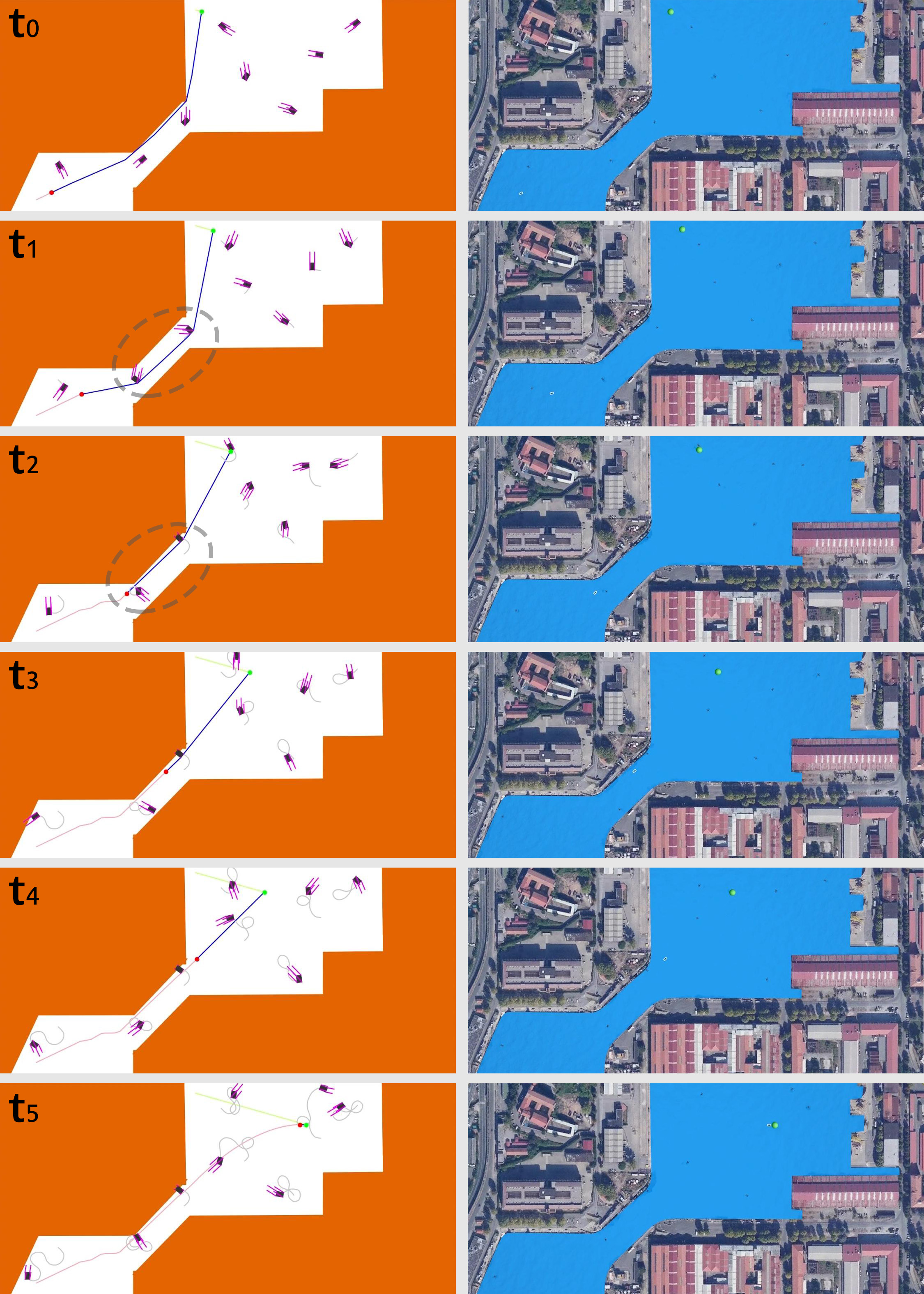}
\caption{\textcolor{black}{Navigation result of OkayPlan in the harbor environment.}}
\label{Harbor_A}
\end{figure*}

\begin{figure*}[h]
\centering
\includegraphics[width=0.88\textwidth]{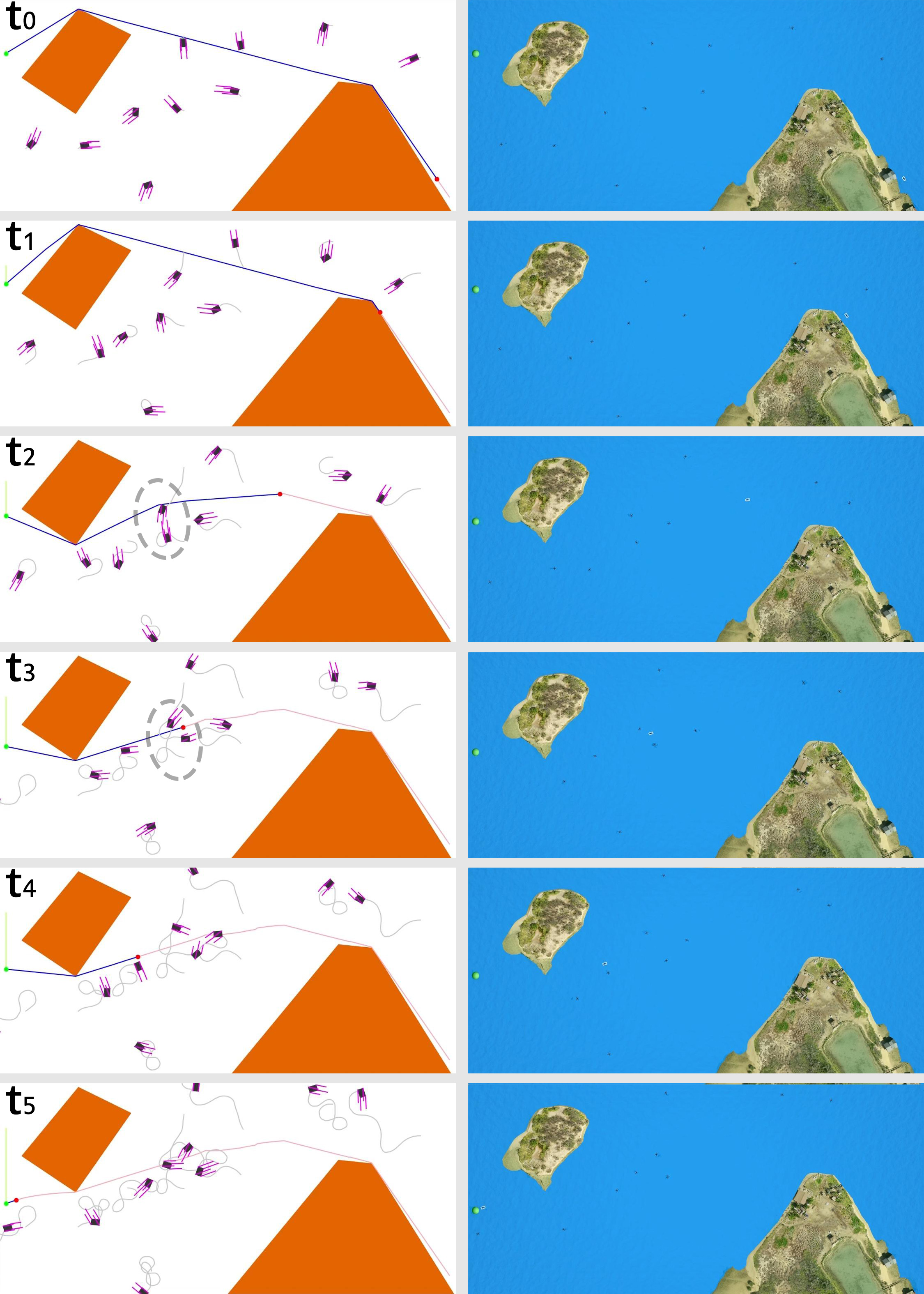}
\caption{\textcolor{black}{Navigation result of OkayPlan in the island environment.}}
\label{Island_B}
\end{figure*}

In this section, we aim to validate the applicability of OkayPlan in more realistic USV navigation scenarios. We employed the VRX simulation platform, which incorporates environmental disturbances as well as the dynamic characteristics of the USV. The dynamic obstacles are constituted by other participating USVs moving randomly in the marine area. The planning results are presented in Fig. \ref{3Dshow}, Fig. \ref{Harbor_A}, and Fig. \ref{Island_B}.

In Fig. \ref{3Dshow} (also see $t_1$ of Fig. \ref{Harbor_A}), the OkayPlan proactively avoids potential collisions by circumventing other USVs. Meanwhile, it regenerates the shortest path once the conflict risk has dissipated, as observed in $t_2$ of Fig. \ref{Harbor_A}. Similar phenomena can be witnessed in $t_2$ and $t_3$ of Fig. \ref{Island_B}. It is noteworthy that such reactive planning capabilities are achieved at a high frequency of 125 Hz, indicating that the proposed algorithm is capable of rapidly identifying the safest or shortest viable path as the environmental conditions evolve. These attributes are particularly beneficial for USVs navigation. Firstly, the superiority in path length is advantageous, given USVs' limited onboard energy resources. Meanwhile, considering the delayed maneuverability of USVs and the dynamic nature of marine environments, the capacity to consider the motion of other obstacles and make timely adjustments to the planned path from a global perspective can significantly bolster navigation safety.

\section{Conclusions}
\textcolor{black}{This paper presents OkayPlan, a real-time GPP algorithm tailored for USVs navigating dynamic marine environments. OkayPlan formulates path planning as an OKAOP to accommodate future obstacle motions. Meanwhile, it employs a DPI mechanism to enhance planning capabilities for dynamic scenarios and introduces a relaxation strategy for autonomous hyperparameter tuning. Comprehensive comparisons with other GPP algorithms demonstrate OkayPlan's superior performance in terms of path safety, path length optimality, and computational efficiency. Notably, OkayPlan achieves impressive real-time planning frequencies of 125 Hz (0.008 \textit{s/p}). Ablation studies further substantiate the crucial contributions of these proposed components. In addition, realistic simulations conducted within the VRX platform solidify its effectiveness in marine environments. Promising future research directions include extending OkayPlan to multi-agent planning scenarios and exploring its applications in real-world USV deployments.}

\section*{Acknowledgement} We acknowledge the support from the National Natural Science Foundation of China under Grant No. 62273230 and 62203302. 

We also acknowledge the support from the Ministry of Education, Singapore, under its Academic Research Fund Tier 1 (RG136/22). Any opinions, findings and conclusions or recommendations expressed in this material are those of the authors and do not reflect the views of the Ministry of Education, Singapore.

\printcredits

\bibliographystyle{cas-model2-names}
\bibliography{OkayPlan}

\begin{thebibliography}{38}
\expandafter\ifx\csname natexlab\endcsname\relax\def\natexlab#1{#1}\fi
\providecommand{\url}[1]{\texttt{#1}}
\providecommand{\href}[2]{#2}
\providecommand{\path}[1]{#1}
\providecommand{\DOIprefix}{doi:}
\providecommand{\ArXivprefix}{arXiv:}
\providecommand{\URLprefix}{URL: }
\providecommand{\Pubmedprefix}{pmid:}
\providecommand{\doi}[1]{\href{http://dx.doi.org/#1}{\path{#1}}}
\providecommand{\Pubmed}[1]{\href{pmid:#1}{\path{#1}}}
\providecommand{\bibinfo}[2]{#2}
\ifx\xfnm\relax \def\xfnm[#1]{\unskip,\space#1}\fi
\bibitem[{Bai et~al.(2023)Bai, Li, Xu and Xiao}]{PGR}
\bibinfo{author}{Bai, X.}, \bibinfo{author}{Li, B.}, \bibinfo{author}{Xu, X.}, \bibinfo{author}{Xiao, Y.}, \bibinfo{year}{2023}.
\newblock \bibinfo{title}{{USV} path planning algorithm based on plant growth}.
\newblock \bibinfo{journal}{Ocean Engineering} \bibinfo{volume}{273}, \bibinfo{pages}{113965}.
\bibitem[{Bayrak and Bayram(2023)}]{VRX_plan}
\bibinfo{author}{Bayrak, M.}, \bibinfo{author}{Bayram, H.}, \bibinfo{year}{2023}.
\newblock \bibinfo{title}{Colreg-compliant simulation environment for verifying {USV} motion planning algorithms}, in: \bibinfo{booktitle}{OCEANS 2023-Limerick}, \bibinfo{organization}{IEEE}. pp. \bibinfo{pages}{1--10}.
\bibitem[{Bingham et~al.(2019)Bingham, Ag{\"u}ero, McCarrin, Klamo, Malia, Allen, Lum, Rawson and Waqar}]{VRX}
\bibinfo{author}{Bingham, B.}, \bibinfo{author}{Ag{\"u}ero, C.}, \bibinfo{author}{McCarrin, M.}, \bibinfo{author}{Klamo, J.}, \bibinfo{author}{Malia, J.}, \bibinfo{author}{Allen, K.}, \bibinfo{author}{Lum, T.}, \bibinfo{author}{Rawson, M.}, \bibinfo{author}{Waqar, R.}, \bibinfo{year}{2019}.
\newblock \bibinfo{title}{Toward maritime robotic simulation in gazebo}, in: \bibinfo{booktitle}{OCEANS 2019 MTS/IEEE SEATTLE}, \bibinfo{organization}{IEEE}. pp. \bibinfo{pages}{1--10}.
\bibitem[{Daniel et~al.(2010)Daniel, Nash, Koenig and Felner}]{theta_star}
\bibinfo{author}{Daniel, K.}, \bibinfo{author}{Nash, A.}, \bibinfo{author}{Koenig, S.}, \bibinfo{author}{Felner, A.}, \bibinfo{year}{2010}.
\newblock \bibinfo{title}{Theta*: Any-angle path planning on grids}.
\newblock \bibinfo{journal}{Journal of Artificial Intelligence Research} \bibinfo{volume}{39}, \bibinfo{pages}{533--579}.
\bibitem[{Dijkstra(1959)}]{Dijkstra}
\bibinfo{author}{Dijkstra, E.}, \bibinfo{year}{1959}.
\newblock \bibinfo{title}{A note on two problems in connexion with graphs}.
\newblock \bibinfo{journal}{Numerische Mathematik} \bibinfo{volume}{1}, \bibinfo{pages}{269--271}.
\bibitem[{Ding et~al.(2023)Ding, Zhou, Huang, Song, Lu and Wang}]{EP_RRTstar}
\bibinfo{author}{Ding, J.}, \bibinfo{author}{Zhou, Y.}, \bibinfo{author}{Huang, X.}, \bibinfo{author}{Song, K.}, \bibinfo{author}{Lu, S.}, \bibinfo{author}{Wang, L.}, \bibinfo{year}{2023}.
\newblock \bibinfo{title}{An improved {RRT*} algorithm for robot path planning based on path expansion heuristic sampling}.
\newblock \bibinfo{journal}{Journal of Computational Science} \bibinfo{volume}{67}, \bibinfo{pages}{101937}.
\bibitem[{Dong et~al.(2023)Dong, Zhang, Qi, Zhang, Li and Liu}]{USV_underact}
\bibinfo{author}{Dong, Z.}, \bibinfo{author}{Zhang, Z.}, \bibinfo{author}{Qi, S.}, \bibinfo{author}{Zhang, H.}, \bibinfo{author}{Li, J.}, \bibinfo{author}{Liu, Y.}, \bibinfo{year}{2023}.
\newblock \bibinfo{title}{Autonomous cooperative formation control of underactuated {USVs} based on improved {MPC} in complex ocean environment}.
\newblock \bibinfo{journal}{Ocean Engineering} \bibinfo{volume}{270}, \bibinfo{pages}{113633}.
\bibitem[{Fu et~al.(2011)Fu, Ding and Zhou}]{PSO_UAV}
\bibinfo{author}{Fu, Y.}, \bibinfo{author}{Ding, M.}, \bibinfo{author}{Zhou, C.}, \bibinfo{year}{2011}.
\newblock \bibinfo{title}{Phase angle-encoded and quantum-behaved particle swarm optimization applied to three-dimensional route planning for {UAV}}.
\newblock \bibinfo{journal}{IEEE Transactions on Systems, Man, and Cybernetics-Part A: Systems and Humans} \bibinfo{volume}{42}, \bibinfo{pages}{511--526}.
\bibitem[{Gammell et~al.(2014)Gammell, Srinivasa and Barfoot}]{IRRTstar}
\bibinfo{author}{Gammell, J.D.}, \bibinfo{author}{Srinivasa, S.S.}, \bibinfo{author}{Barfoot, T.D.}, \bibinfo{year}{2014}.
\newblock \bibinfo{title}{Informed {RRT*}: Optimal sampling-based path planning focused via direct sampling of an admissible ellipsoidal heuristic}, in: \bibinfo{booktitle}{2014 IEEE/RSJ international conference on intelligent robots and systems}, \bibinfo{organization}{IEEE}. pp. \bibinfo{pages}{2997--3004}.
\bibitem[{Gonzalez{-}Garcia and Castaneda(2021)}]{USV_uncertainties}
\bibinfo{author}{Gonzalez{-}Garcia, A.}, \bibinfo{author}{Castaneda, H.}, \bibinfo{year}{2021}.
\newblock \bibinfo{title}{Guidance and control based on adaptive sliding mode strategy for a {USV} subject to uncertainties}.
\newblock \bibinfo{journal}{IEEE Journal of Oceanic Engineering} \bibinfo{volume}{46}, \bibinfo{pages}{1144--1154}.
\bibitem[{Guo et~al.(2020a)Guo, Ji, Zhao, Wen and Zhang}]{PSO_USV}
\bibinfo{author}{Guo, X.}, \bibinfo{author}{Ji, M.}, \bibinfo{author}{Zhao, Z.}, \bibinfo{author}{Wen, D.}, \bibinfo{author}{Zhang, W.}, \bibinfo{year}{2020}a.
\newblock \bibinfo{title}{Global path planning and multi-objective path control for unmanned surface vehicle based on modified particle swarm optimization ({PSO}) algorithm}.
\newblock \bibinfo{journal}{Ocean Engineering} \bibinfo{volume}{216}, \bibinfo{pages}{107693}.
\bibitem[{Guo et~al.(2020b)Guo, Ji, Zhao, Wen and Zhang}]{IPSO1}
\bibinfo{author}{Guo, X.}, \bibinfo{author}{Ji, M.}, \bibinfo{author}{Zhao, Z.}, \bibinfo{author}{Wen, D.}, \bibinfo{author}{Zhang, W.}, \bibinfo{year}{2020}b.
\newblock \bibinfo{title}{Global path planning and multi-objective path control for unmanned surface vehicle based on modified particle swarm optimization ({PSO}) algorithm}.
\newblock \bibinfo{journal}{Ocean Engineering} \bibinfo{volume}{216}, \bibinfo{pages}{107693}.
\bibitem[{Harabor and Grastien(2011)}]{JPS}
\bibinfo{author}{Harabor, D.}, \bibinfo{author}{Grastien, A.}, \bibinfo{year}{2011}.
\newblock \bibinfo{title}{Online graph pruning for pathfinding on grid maps}, in: \bibinfo{booktitle}{Proceedings of the AAAI conference on artificial intelligence}, pp. \bibinfo{pages}{1114--1119}.
\bibitem[{Hart et~al.(1968)Hart, Nilsson and Raphael}]{Astar}
\bibinfo{author}{Hart, P.E.}, \bibinfo{author}{Nilsson, N.J.}, \bibinfo{author}{Raphael, B.}, \bibinfo{year}{1968}.
\newblock \bibinfo{title}{A formal basis for the heuristic determination of minimum cost paths}.
\newblock \bibinfo{journal}{IEEE Transactions on Systems Science and Cybernetics} \bibinfo{volume}{4}, \bibinfo{pages}{100--107}.
\newblock \DOIprefix\doi{10.1109/TSSC.1968.300136}.
\bibitem[{Holland(1992)}]{GA}
\bibinfo{author}{Holland, J.H.}, \bibinfo{year}{1992}.
\newblock \bibinfo{title}{Adaptation in natural and artificial systems: an introductory analysis with applications to biology, control, and artificial intelligence}.
\newblock \bibinfo{publisher}{MIT press}.
\bibitem[{Hu et~al.(2023)Hu, Wang, Cheng and Gao}]{predict1}
\bibinfo{author}{Hu, H.}, \bibinfo{author}{Wang, Q.}, \bibinfo{author}{Cheng, M.}, \bibinfo{author}{Gao, Z.}, \bibinfo{year}{2023}.
\newblock \bibinfo{title}{Trajectory prediction neural network and model interpretation based on temporal pattern attention}.
\newblock \bibinfo{journal}{IEEE Transactions on Intelligent Transportation Systems} \bibinfo{volume}{24}, \bibinfo{pages}{2746--2759}.
\newblock \DOIprefix\doi{10.1109/TITS.2022.3219874}.
\bibitem[{Islam et~al.(2012)Islam, Nasir, Malik, Ayaz and Hasan}]{RRTstar_smart}
\bibinfo{author}{Islam, F.}, \bibinfo{author}{Nasir, J.}, \bibinfo{author}{Malik, U.}, \bibinfo{author}{Ayaz, Y.}, \bibinfo{author}{Hasan, O.}, \bibinfo{year}{2012}.
\newblock \bibinfo{title}{{RRT*}-smart: Rapid convergence implementation of {RRT*} towards optimal solution}, in: \bibinfo{booktitle}{2012 IEEE international conference on mechatronics and automation}, \bibinfo{organization}{IEEE}. pp. \bibinfo{pages}{1651--1656}.
\bibitem[{Jin et~al.(2018)Jin, Zhang, Shao, Lyu and Wang}]{app_ore}
\bibinfo{author}{Jin, J.}, \bibinfo{author}{Zhang, J.}, \bibinfo{author}{Shao, F.}, \bibinfo{author}{Lyu, Z.}, \bibinfo{author}{Wang, D.}, \bibinfo{year}{2018}.
\newblock \bibinfo{title}{A novel ocean bathymetry technology based on an unmanned surface vehicle}.
\newblock \bibinfo{journal}{Acta Oceanologica Sinica} \bibinfo{volume}{37}, \bibinfo{pages}{99--106}.
\bibitem[{Kaindl and Kainz(1997)}]{BAstar}
\bibinfo{author}{Kaindl, H.}, \bibinfo{author}{Kainz, G.}, \bibinfo{year}{1997}.
\newblock \bibinfo{title}{Bidirectional heuristic search reconsidered}.
\newblock \bibinfo{journal}{Journal of Artificial Intelligence Research} \bibinfo{volume}{7}, \bibinfo{pages}{283--317}.
\bibitem[{Karaman and Frazzoli(2011)}]{RRTstar}
\bibinfo{author}{Karaman, S.}, \bibinfo{author}{Frazzoli, E.}, \bibinfo{year}{2011}.
\newblock \bibinfo{title}{Sampling-based algorithms for optimal motion planning}.
\newblock \bibinfo{journal}{The international journal of robotics research} \bibinfo{volume}{30}, \bibinfo{pages}{846--894}.
\bibitem[{Kavraki et~al.(1996)Kavraki, Svestka, Latombe and Overmars}]{PRM}
\bibinfo{author}{Kavraki, L.E.}, \bibinfo{author}{Svestka, P.}, \bibinfo{author}{Latombe, J.C.}, \bibinfo{author}{Overmars, M.H.}, \bibinfo{year}{1996}.
\newblock \bibinfo{title}{Probabilistic roadmaps for path planning in high-dimensional configuration spaces}.
\newblock \bibinfo{journal}{IEEE transactions on Robotics and Automation} \bibinfo{volume}{12}, \bibinfo{pages}{566--580}.
\bibitem[{Kennedy and Eberhart(1995)}]{PSO}
\bibinfo{author}{Kennedy, J.}, \bibinfo{author}{Eberhart, R.}, \bibinfo{year}{1995}.
\newblock \bibinfo{title}{Particle swarm optimization}, in: \bibinfo{booktitle}{Proceedings of ICNN'95-international conference on neural networks}, \bibinfo{organization}{IEEE}. pp. \bibinfo{pages}{1942--1948}.
\bibitem[{Kim et~al.(2023)Kim, Jeon, Choi and Kum}]{predict2}
\bibinfo{author}{Kim, S.}, \bibinfo{author}{Jeon, H.}, \bibinfo{author}{Choi, J.W.}, \bibinfo{author}{Kum, D.}, \bibinfo{year}{2023}.
\newblock \bibinfo{title}{Diverse multiple trajectory prediction using a two-stage prediction network trained with lane loss}.
\newblock \bibinfo{journal}{IEEE Robotics and Automation Letters} \bibinfo{volume}{8}, \bibinfo{pages}{2038--2045}.
\newblock \DOIprefix\doi{10.1109/LRA.2022.3231525}.
\bibitem[{LaValle(1998)}]{RRT}
\bibinfo{author}{LaValle, S.}, \bibinfo{year}{1998}.
\newblock \bibinfo{title}{Rapidly-exploring random trees: A new tool for path planning}.
\newblock \bibinfo{journal}{Research Report 9811} .
\bibitem[{Li et~al.(2018)Li, Wang and Du}]{app_ep}
\bibinfo{author}{Li, D.}, \bibinfo{author}{Wang, P.}, \bibinfo{author}{Du, L.}, \bibinfo{year}{2018}.
\newblock \bibinfo{title}{Path planning technologies for autonomous underwater vehicles-a review}.
\newblock \bibinfo{journal}{Ieee Access} \bibinfo{volume}{7}, \bibinfo{pages}{9745--9768}.
\bibitem[{Li and Chou(2018)}]{PSO_mobilerobot}
\bibinfo{author}{Li, G.}, \bibinfo{author}{Chou, W.}, \bibinfo{year}{2018}.
\newblock \bibinfo{title}{Path planning for mobile robot using self-adaptive learning particle swarm optimization}.
\newblock \bibinfo{journal}{Science China Information Sciences} \bibinfo{volume}{61}, \bibinfo{pages}{1--18}.
\bibitem[{Li and Yu(2023)}]{IPSO2}
\bibinfo{author}{Li, X.}, \bibinfo{author}{Yu, S.}, \bibinfo{year}{2023}.
\newblock \bibinfo{title}{Three-dimensional path planning for {AUVs} in ocean currents environment based on an improved compression factor particle swarm optimization algorithm}.
\newblock \bibinfo{journal}{Ocean Engineering} \bibinfo{volume}{280}, \bibinfo{pages}{114610}.
\bibitem[{Liu et~al.(2016)Liu, Zhang, Yu and Yuan}]{app_mtm}
\bibinfo{author}{Liu, Z.}, \bibinfo{author}{Zhang, Y.}, \bibinfo{author}{Yu, X.}, \bibinfo{author}{Yuan, C.}, \bibinfo{year}{2016}.
\newblock \bibinfo{title}{Unmanned surface vehicles: An overview of developments and challenges}.
\newblock \bibinfo{journal}{Annual Reviews in Control} \bibinfo{volume}{41}, \bibinfo{pages}{71--93}.
\bibitem[{Moore(1959)}]{BFS}
\bibinfo{author}{Moore, E.F.}, \bibinfo{year}{1959}.
\newblock \bibinfo{title}{The shortest path through a maze}, in: \bibinfo{booktitle}{Proc. of the International Symposium on the Theory of Switching}, \bibinfo{organization}{Harvard University Press}. pp. \bibinfo{pages}{285--292}.
\bibitem[{Roberge et~al.(2012)Roberge, Tarbouchi and Labont{\'e}}]{PGA}
\bibinfo{author}{Roberge, V.}, \bibinfo{author}{Tarbouchi, M.}, \bibinfo{author}{Labont{\'e}, G.}, \bibinfo{year}{2012}.
\newblock \bibinfo{title}{Comparison of parallel genetic algorithm and particle swarm optimization for real-time {UAV} path planning}.
\newblock \bibinfo{journal}{IEEE Transactions on industrial informatics} \bibinfo{volume}{9}, \bibinfo{pages}{132--141}.
\bibitem[{Schofield et~al.(2018)Schofield, Wilde and Murphy}]{app_wr}
\bibinfo{author}{Schofield, R.T.}, \bibinfo{author}{Wilde, G.A.}, \bibinfo{author}{Murphy, R.R.}, \bibinfo{year}{2018}.
\newblock \bibinfo{title}{Potential field implementation for move-to-victim behavior for a lifeguard assistant unmanned surface vehicle}, in: \bibinfo{booktitle}{2018 IEEE International Symposium on Safety, Security, and Rescue Robotics (SSRR)}, pp. \bibinfo{pages}{1--2}.
\newblock \DOIprefix\doi{10.1109/SSRR.2018.8468602}.
\bibitem[{Wen et~al.(2020)Wen, Tao, Zhu, Zhou and Xiao}]{USV_nonlinear}
\bibinfo{author}{Wen, Y.}, \bibinfo{author}{Tao, W.}, \bibinfo{author}{Zhu, M.}, \bibinfo{author}{Zhou, J.}, \bibinfo{author}{Xiao, C.}, \bibinfo{year}{2020}.
\newblock \bibinfo{title}{Characteristic model-based path following controller design for the unmanned surface vessel}.
\newblock \bibinfo{journal}{Applied Ocean Research} \bibinfo{volume}{101}, \bibinfo{pages}{102293}.
\bibitem[{Wu and Wei(2023)}]{VRX_ctrl}
\bibinfo{author}{Wu, X.}, \bibinfo{author}{Wei, C.}, \bibinfo{year}{2023}.
\newblock \bibinfo{title}{{DRL-Based} motion control for unmanned surface vehicles with environmental disturbances}, in: \bibinfo{booktitle}{2023 IEEE International Conference on Unmanned Systems (ICUS)}, pp. \bibinfo{pages}{1696--1700}.
\newblock \DOIprefix\doi{10.1109/ICUS58632.2023.10318284}.
\bibitem[{Xin et~al.(2024)Xin, Li, Zhang and Li}]{SEPSO}
\bibinfo{author}{Xin, J.}, \bibinfo{author}{Li, Z.}, \bibinfo{author}{Zhang, Y.}, \bibinfo{author}{Li, N.}, \bibinfo{year}{2024}.
\newblock \bibinfo{title}{Efficient real-time path planning with self-evolving particle swarm optimization in dynamic scenarios}.
\newblock \bibinfo{journal}{Unmanned Systems} \bibinfo{volume}{12}, \bibinfo{pages}{215--226}.
\newblock \DOIprefix\doi{10.1142/S230138502441005X}.
\bibitem[{Xin et~al.(2023)Xin, Yu, Wang and Li}]{DPPSO}
\bibinfo{author}{Xin, J.}, \bibinfo{author}{Yu, L.}, \bibinfo{author}{Wang, J.}, \bibinfo{author}{Li, N.}, \bibinfo{year}{2023}.
\newblock \bibinfo{title}{A diversity-based parallel particle swarm optimization for nonconvex economic dispatch problem}.
\newblock \bibinfo{journal}{Transactions of the Institute of Measurement and Control} \bibinfo{volume}{45}, \bibinfo{pages}{452--465}.
\bibitem[{Xu et~al.(2024a)Xu, Yang, Zhou and Xu}]{OE_USV_IBA}
\bibinfo{author}{Xu, D.}, \bibinfo{author}{Yang, J.}, \bibinfo{author}{Zhou, X.}, \bibinfo{author}{Xu, H.}, \bibinfo{year}{2024}a.
\newblock \bibinfo{title}{Hybrid path planning method for {USV} using bidirectional {A*} and improved {DWA} considering the manoeuvrability and {COLREGs}}.
\newblock \bibinfo{journal}{Ocean Engineering} \bibinfo{volume}{298}, \bibinfo{pages}{117210}.
\bibitem[{Xu et~al.(2024b)Xu, Cao, Cai, Zhang and Chen}]{OE_USV_DDPG}
\bibinfo{author}{Xu, X.}, \bibinfo{author}{Cao, Y.}, \bibinfo{author}{Cai, P.}, \bibinfo{author}{Zhang, W.}, \bibinfo{author}{Chen, H.}, \bibinfo{year}{2024}b.
\newblock \bibinfo{title}{Research on real-time collision avoidance and path planning of {USVs} in multi-obstacle ships environment}.
\newblock \bibinfo{journal}{Ocean Engineering} \bibinfo{volume}{295}, \bibinfo{pages}{116890}.
\bibitem[{Yao et~al.(2023)Yao, Lou and Zhang}]{OE_MultiUSV}
\bibinfo{author}{Yao, P.}, \bibinfo{author}{Lou, Y.}, \bibinfo{author}{Zhang, K.}, \bibinfo{year}{2023}.
\newblock \bibinfo{title}{Multi-{USV} cooperative path planning by window update based self-organizing map and spectral clustering}.
\newblock \bibinfo{journal}{Ocean Engineering} \bibinfo{volume}{275}, \bibinfo{pages}{114140}.

\end{thebibliography}

\clearpage
\onecolumn
\section*{Appendix}

\twocolumn

\end{document}